\documentclass[12pt]{iopart}

\usepackage{harvard}

\usepackage{hyperref}
\usepackage{lipsum}
\usepackage{graphicx}
\usepackage{subfigure}
\usepackage{multirow}
\usepackage{algorithmicx,algorithm}
\usepackage{xcolor}
\usepackage{amsfonts}
\usepackage{amssymb}
\expandafter\let\csname equation*\endcsname\relax
\expandafter\let\csname endequation*\endcsname\relax
\usepackage{amsmath}
\usepackage{color}
\usepackage[T1]{fontenc}
\usepackage{siunitx}

\newcommand\blfootnote[1]{%
	\begingroup
	\renewcommand\thefootnote{}\footnote{#1}%
	\addtocounter{footnote}{-1}%
	\endgroup
}

\begin{document}

\title{Intensive Vision-guided Network for Radiology Report Generation}

\author{Fudan Zheng$^1$, Mengfei Li$^1$\footnote{Equal contribution as Fudan Zheng}, Ying Wang$^2$, Weijiang Yu$^3$\blfootnote{Corresponding author: weijiangyu8@gmail.com}, Ruixuan Wang$^1$, Zhiguang Chen$^{1,2}$, Nong Xiao$^{1,2}$, and Yutong Lu$^{1,2}$\blfootnote{Corresponding author: yutong.lu@nscc-gz.cn}}

\address{$^1$Sun Yat-Sen University, No. 132 Waihuandong Road, Guangzhou Higher Education Mega Center, Guangzhou, 510006, China}
\address{$^2$National SuperComputer Center in Guangzhou, No. 132 Waihuandong Road, Guangzhou Higher Education Mega Center, Guangzhou, 510006, China}
\address{$^3$Huawei Technologies Co., Ltd., Huawei Industrial Park, Bantian, Longgang District, Shenzhen, 518129, China}
\ead{weijiangyu8@gmail.com;yutong.lu@nscc-gz.cn}

\vspace{10pt}


\begin{abstract}
\textit{Objective.} Automatic radiology report generation is booming due to its huge application potential for the healthcare industry. However, existing computer vision and natural language processing approaches to tackle this problem are limited in two aspects. First, when extracting image features, most of them neglect multi-view reasoning in vision and model single-view structure of medical images, such as space-view or channel-view. However, clinicians rely on multi-view imaging information for comprehensive judgment in daily clinical diagnosis. Second, when generating reports, they overlook context reasoning with multi-modal information and focus on pure textual optimization utilizing retrieval-based methods. We aim to address these two issues by proposing a model that better simulates clinicians’ perspectives and generates more accurate reports. \textit{Approach.} Given the above limitation in feature extraction, we propose a Globally-intensive Attention (GIA) module in the medical image encoder to simulate and integrate multi-view vision perception. GIA aims to learn three types of vision perception: depth view, space view, and pixel view. On the other hand, to address the above problem in report generation, we explore how to involve multi-modal signals to generate precisely matched reports, i.e., how to integrate previously predicted words with region-aware visual content in next word prediction. Specifically, we design a Visual Knowledge-guided Decoder (VKGD), which can adaptively consider how much the model needs to rely on visual information and previously predicted text to assist next word prediction. Hence, our final Intensive Vision-guided Network (IVGN) framework includes a GIA-guided Visual Encoder and the VKGD. \textit{Main results.} Experiments on two commonly-used datasets \textsc{IU X-Ray} and MIMIC-CXR demonstrate the superior ability of our method compared with other state-of-the-art approaches. \textit{Significance.} Our model explores the potential of simulating clinicians’ perspectives and automatically generates more accurate reports, which promotes the exploration of medical automation and intelligence.

\end{abstract}

%

\submitto{\PMB}
%
%
%

\section{Introduction}
\label{sec:introduction}

Medical imaging has become a very important examination way in clinical medicine. In clinical diagnosis, disease classification, lesion detection, medical segmentation or prognosis prediction are often carried out based on patients' examined radiology images, such as X-ray, CT, MRI, etc \cite{C1,C2,C3,C4}.  Radiologists usually need to write radiology reports based on the examined images to provide evidence for patients' final diagnosis. Writing these reports is time-consuming and labor-intensive. On the one hand, daily inspections bring about a huge amount of imaging data to be dealt with. On the other hand, the reports shall contain a variety of normal and abnormal medical observations, which are cumbersome to write. Therefore, automatic generation of radiology report can greatly reduce the burden of radiologists and thus promote the development of medical automation and intelligence.

\begin{figure}[!t]
	\centerline{\includegraphics[width=0.7\columnwidth]{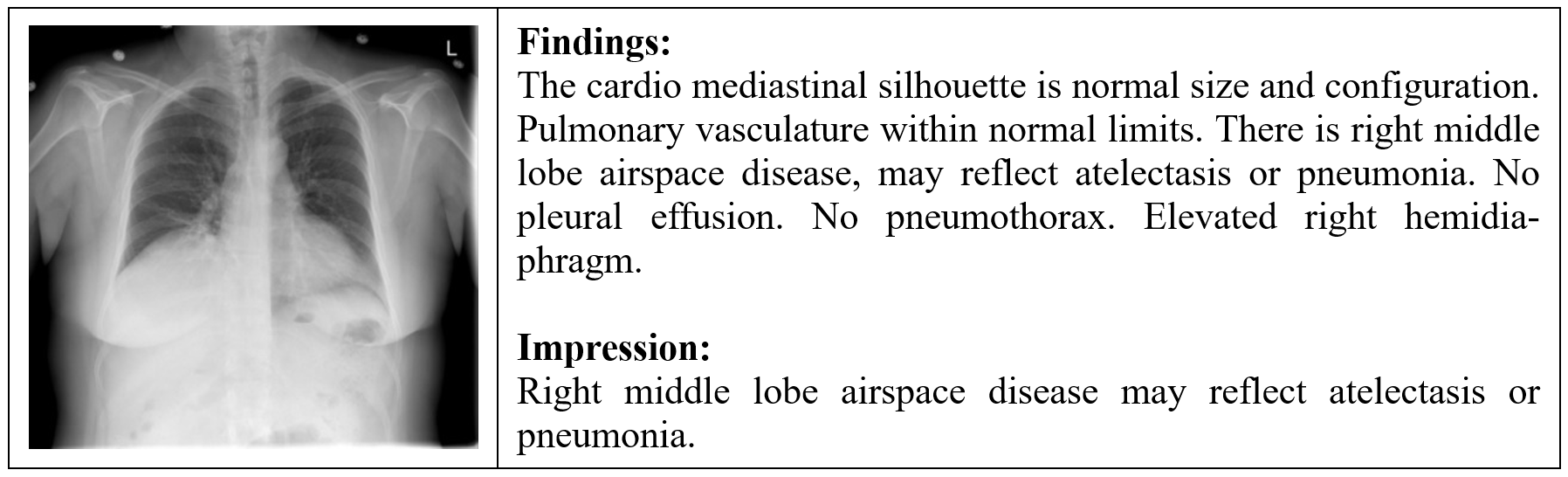}}
	\caption{An example of chest X-ray image and its report. The task of automatic radiology report generation is to automatically generate a report based on the given radiology image.}
	\label{fig_1}
\end{figure}

In recent years, automatic generation of radiology reports has become a popular research topic due to its huge application potential \cite{C1,C3,C5,C6,C7,C8,C9,C10,C11,C12,C13,C14,C15,C16}. It aims to automatically generate a report that clearly describes the normal and abnormal medical findings based on a radiology image.  Take chest X-ray image as an example, as shown in Figure \ref{fig_1}, the radiology report contains the \emph{Findings} section which describes medical findings in detail, including normal and abnormal characteristics, and the \emph{Impression} section that summarizes the most important and significant findings or conclusions. With the development of deep visual learning and natural language processing, many studies apply the most advanced technologies in these two fields to automatic generation of radiology reports \cite{C5,C6,C7,C8,C9,C10,C11,C12,C13,C14,C15,C16}. However, these approaches do not completely address the following two challenges inherent in the task.

\begin{figure}[!t]
	\centerline{\includegraphics[width=0.7\columnwidth]{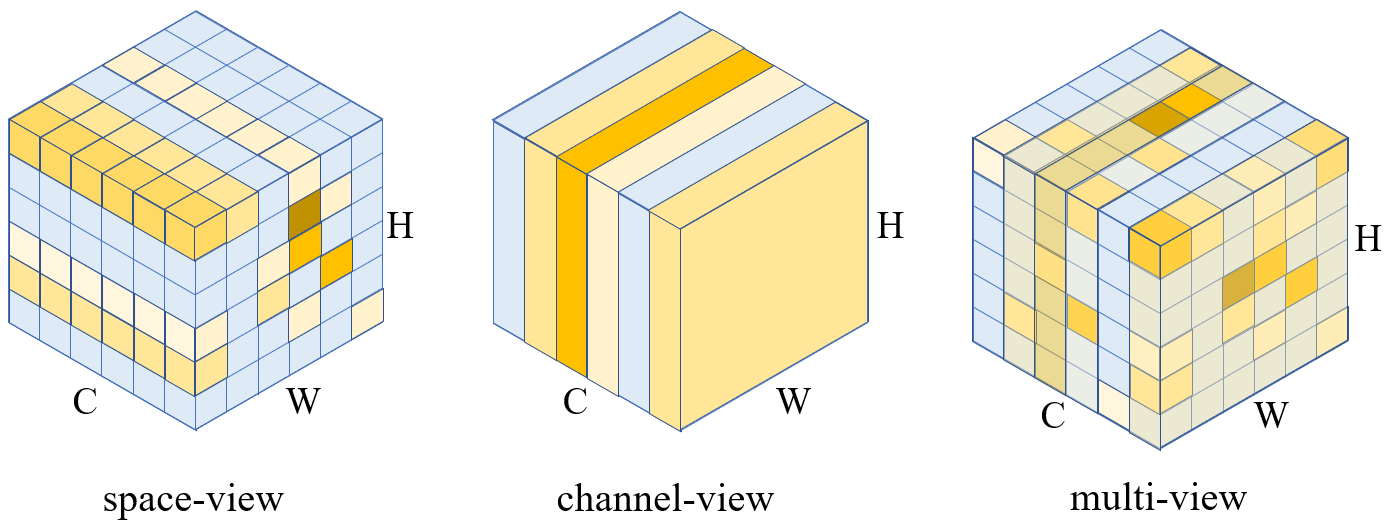}}
	\caption{A schematic diagram of attention mechanism from different views. The left and middle subgraphs represent attention modeling only from the space view and channel view, respectively. The model can only learn the importance of different positions in the same space or the importance of different channels. The subgraph on the right represents multi-view attention modeling, in which the model can learn the importance of each position in the feature maps and ultimately guide the model to learn the most salient visual features. Best viewed in color.}
	\label{fig_2}
\end{figure}

First, vision analysis of radiology images needs to be carried out from multiple perspectives. In daily clinical diagnosis, clinicians will examine patients' radiology images from multiple views, especially when certain physiological parts or organs are partially obscured or totally out of sight at certain views. These multi-view imaging information are very important for clinicians to make a comprehensive  diagnosis of the radiology images. However, in visual learning of medical images, previous studies usually use CNN for feature extraction \cite{C7,C8,C10,C11} and employ space or channel attention mechanisms to endow more important visual regions with greater focus \cite{C6}, or use the recently proposed Transformer architecture to learn long-range relationship between various patches \cite{C5,C14,C15}. They mainly focus on single-view structure modeling of medical images, such as space view or channel view, and neglect multi-view reasoning in vision. Figure \ref{fig_2} shows the differences of attention mechanism from different views. The left and middle subgraphs represent attention modeling only from the space view and channel view, respectively. The model can only learn the importance of different positions in the same space or the importance of different channels. The subgraph on the right represents multi-view attention modeling, in which the model can learn the importance of each position in the feature maps and ultimately guide the model to learn the most salient visual features. Therefore, on vision analysis, our motivation is to simulate clinicians' multi-view vision perception. To this end, we propose a Globally-intensive Attention (GIA) module in the visual encoder. Specifically, we propose a depth-view Batch Normalization-guided Weight Adapter (BNWA), a space-view BNWA and a pixel-view attention mechanism to mine the importance of each position in multi-view feature maps, which ultimately guide the model to learn the most salient visual features.

\begin{figure}[!t]
	\centerline{\includegraphics[width=1\columnwidth]{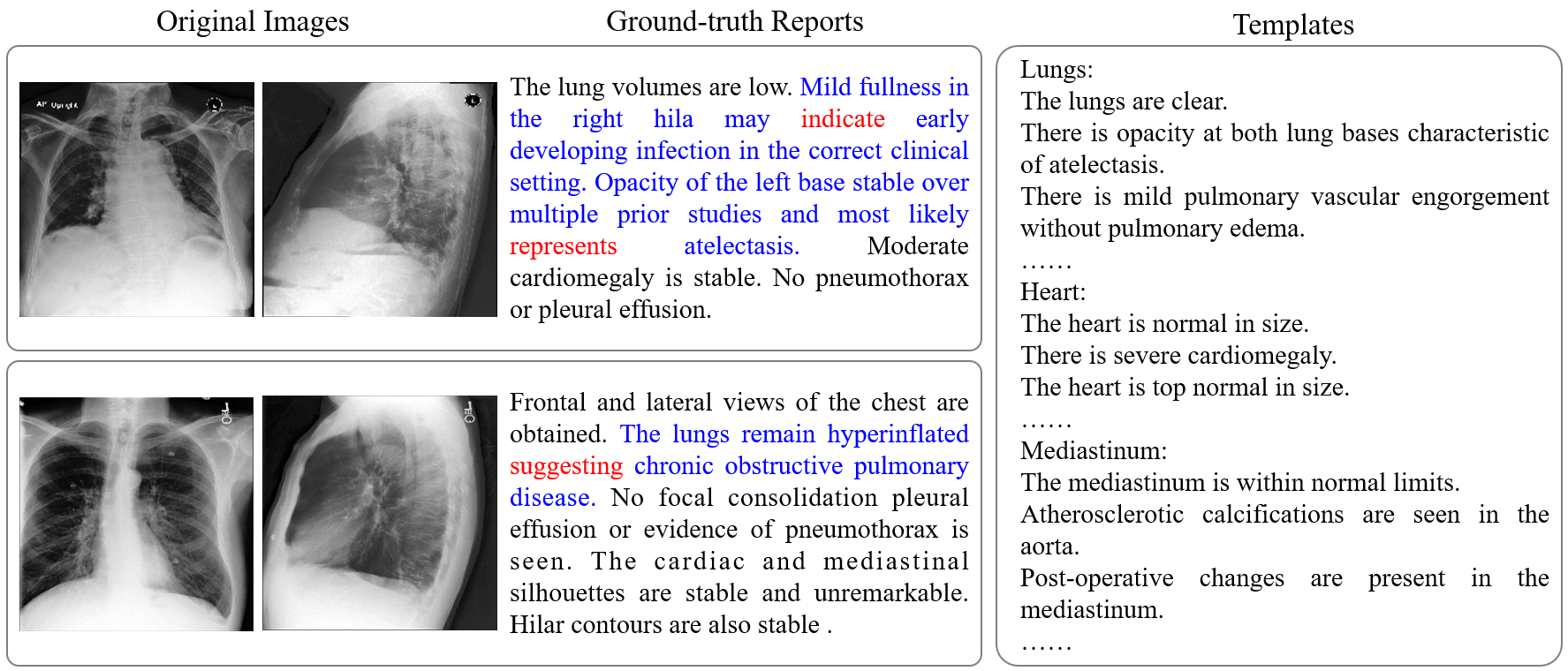}}
	\caption{Examples of ground-truth reports and templates. It can be seen that when writing radiology reports, clinicians often associate the symptoms they find with diseases, and clearly describe “what diseases are derived from what symptoms”, as shown in the blue sentences and red words. However, templates are often limited to specific locations and descriptions and fail to reveal a causal relationship between the abnormalities found and the diseases. Best viewed in color.}
	\label{fig_3}
\end{figure}

Second, different from ordinary text generation task, radiology report generation has the following characteristics: 1) Radiology reports are long narratives composed of multiple sentences, so the logic between sentences is very important. 2) There are higher requirements on professional accuracy, i.e. clinical accuracy for radiology report generation. 3) In radiology reports, disease features are always closely related to their corresponding physiological parts. These characteristics, especially the third one, indicate that the model needs to consider the influence of the previously generated text and the related parts of the image when generating the next word. For example, suppose there are symptoms of emphysema in a radiology image. Then, when the word ``emphysema'' needs to be generated in the report, the word ``lung'' or ``pulmonary'' (rather than the words for other organs) in the previously generated text, as well as the lung area (rather than other organ areas) in the image, becomes very important. However, previous studies tend to focus on pure textual optimization and overlook context reasoning with this multi-modal information. They mostly adopt retrieval-based \cite{C7} or knowledge-based \cite{C10,C11} approaches, which require to retrieve a large template database or to explicitly construct a template list. It can be seen from the ground-truth reports in Figure \ref{fig_3} that when writing radiology reports, clinicians often associate the symptoms they find with diseases, and clearly describe ``what diseases are derived from what symptoms'', as shown in the blue sentences and red words in Figure \ref{fig_3}. However, as can be seen from the templates in Figure \ref{fig_3}, templates are often limited to specific locations and descriptions and fail to reveal a causal relationship between the abnormalities found and the diseases. Therefore, template-based or retrieval-based approaches can neither highlight the most prominent abnormal observations nor generate logical associations between clinical observations and the corresponding diseases in reports. In addition, for the task of radiology report generation, we need to make full use of the extracted image features and the  previously predicted text to guide next word generation, but at the same time, we should not  force the visual attention to be active for every generated word. For example, the text generator likely requires little and even no visual information from the image to predict non-visual words such as ``a'', ``and'' and other conjunctions, etc. Only the generation of words that are really visually relevant needs to rely on the guidance of visual information. Hence, on text generation, our motivation is to design a module that can identify those interdependent parts of the generated text, as well as the really relevant parts of the text and images, and focus on the  clinical  accuracy of the words in the generated reports. To this end, we propose a Visual Knowledge-guided Decoder (VKGD) to adaptively consider how much it needs to rely on visual information and previously predicted text to assist next word generation, rather than completely ignoring visual information or obsessively relying on visual information.

In summary, our work has three main contributions:

1) We propose a novel Intensive Vision-guided Network (IVGN) for automatic radiology report generation, which contains a Globally-intensive Attention (GIA) module in the visual encoder and a Visual Knowledge-guided Decoder (VKGD) for text generation.

2) The GIA module integrates multi-view vision perception of medical images to enhance feature representation. The VKGD can adaptively consider how much it needs to rely on visual information and previously predicted text to assist next word generation, so as to generate more accurate reports.

3) Extensive experiments on two commonly-used benchmark datasets \textsc{IU X-Ray} and MIMIC-CXR demonstrate the superior ability of our proposed Intensive Vision-guided Network (IVGN) to achieve state-of-the-art performance compared with other approaches.

\section{Related Work}

\subsection{Image captioning}

The task of image captioning is to generate readable, accurate and linguistically correct captions for a given image, which is similar to that of our study. The challenge is that the model must not only detect the objects in the image, but also understand how they relate to each other, and finally express them in reasonable language. Most existing image captioning models are based on the CNN-RNN architecture, encoding image features and generating statements separately \cite{C18,C17,C19}. Recent breakthroughs based on attention mechanisms \cite{C18,C17,C20,C21} allow the above image encoding and language modeling steps to be fused together. Reinforcement learning strategies are also employed to better learn the associations between image features and text embedding \cite{C20,C22,C19}, which greatly improve the performance of this task. However, different from image captioning, the automatic radiology report generation task in this study generally needs to generate a longer text, that is, a long narrative consisting of multiple sentences. More importantly, the accuracy of radiology reports will greatly affect the diagnosis of patients, so there are higher requirements on accuracy than other metrics such as smoothness, fluency or elegance for radiology reports. Therefore, models with good performance on image captioning may not be applicable to radiology report generation.

There are some other studies on long text generation \cite{C23}, sentence generation \cite{C24}, paragraph generation \cite{C25}, visual description \cite{C26}, and text summarization \cite{C27}. Some studies achieve good performance in text description of natural images by using GAN \cite{C25}, template-guided module \cite{C28}, attention mechanism \cite{C29} and reinforcement learning \cite{C30}. However, due to the huge difference in pixel value distribution between natural and radiology images, these models may not be suitable for automatic radiology report generation. In contrast, we propose to improve the model in both visual pattern learning and text generation according to the characteristics of radiology images and radiology reports, respectively.

\subsection{Automatic radiology report generation}

The application of deep learning technology to automatic text generation in medical images is a research hotspot. Most previous studies aim at generating fully structured or semi-structured text annotations for medical images, which include predicting attributes of medical images \cite{C31}, creating or predicting labels (location, severity, etc.) for medical images \cite{C32,C33} and generating semi-structured pathological reports limited to certain topics \cite{D1}.

With the development of medical automation and intelligence, more and more studies focus on the automatic generation of unstructured and free-text radiology reports for medical images. Most of them are still based on CNN-RNN architecture, and different strategies, such as reinforcement learning, knowledge graph learning and attention mechanism, are introduced to guide and assist model training.

\textbf{Reinforcement learning.} \citename{C7}\citeyear{C7} put forward a hybrid retrieval generation reinforced agent based on reinforcement learning to learn a report generator that can decide whether to retrieve existing medical templates or to generate new sentences. \citename{C9}\citeyear{C9} focus on modeling the relationship between the Findings section and the Impression section in image reports, and train the whole model via reinforce algorithm. However, both of them use traditional NLP metrics as rewards, rather than using domain-specific rewards that explicitly promote medical accuracy.

\textbf{Knowledge graph learning.} \citename{C12}\citeyear{C12} incorporate a knowledge graph pre-constructed on multiple disease findings to assist feature learning of each disease and the relationship between them. Similarly, \citename{C13}\citeyear{C13} model the associations among medical findings as a knowledge graph and use it to help report generation. \citename{C14}\citeyear{C14} put forward a posterior and prior knowledge based approach, which first examines the abnormal regions and assigns disease tags to the regions, and then writes reports based on years of accumulated medical knowledge and working experience. \citename{C10}\citeyear{C10} combine traditional knowledge-based and retrieval-based methods with modern learning-based methods, formulating medical report writing as a knowledge-driven encode, retrieve and paraphrase process. Ablation studies in the above work show that the involvement of domain expert knowledge does contribute greatly to model performance. However, additional special steps are required to explicitly incorporate domain expert knowledge into model design.

\textbf{Attention mechanism.} \citename{C6}\citeyear{C6} construct a multi-task learning framework which contains a co-attention module and a hierarchical LSTM module to better generate paragraphs of medical reports. \citename{C11}\citeyear{C11} introduce multi-level attention modules into an end-to-end trainable CNN-RNN architecture for considering the meaningful words and image regions. However, at visual level, their attention mechanism is only applied to spatial dimension, lacking multi-view visual reasoning.

In addition, \citename{C5}\citeyear{C5,C16} introduce a memory mechanism into the encoder-decoder framework of Transformer to better depict local structure in image report while modeling global information and to better record the aligment between images and texts respectively. Transformer is based on the self-attention mechanism, which is more effective than the traditional CNN model in mining and processing long-distance dependencies and more sensitive to the location information of objects. Therefore, the framework indeed achieves considerable performance improvement. However, the memory module also brings non-negligible parameters and computational costs.

\section{Method}
\subsection{Overall architecture}

\begin{figure*}[!t]
	\centerline{\includegraphics[width=1.0\columnwidth]{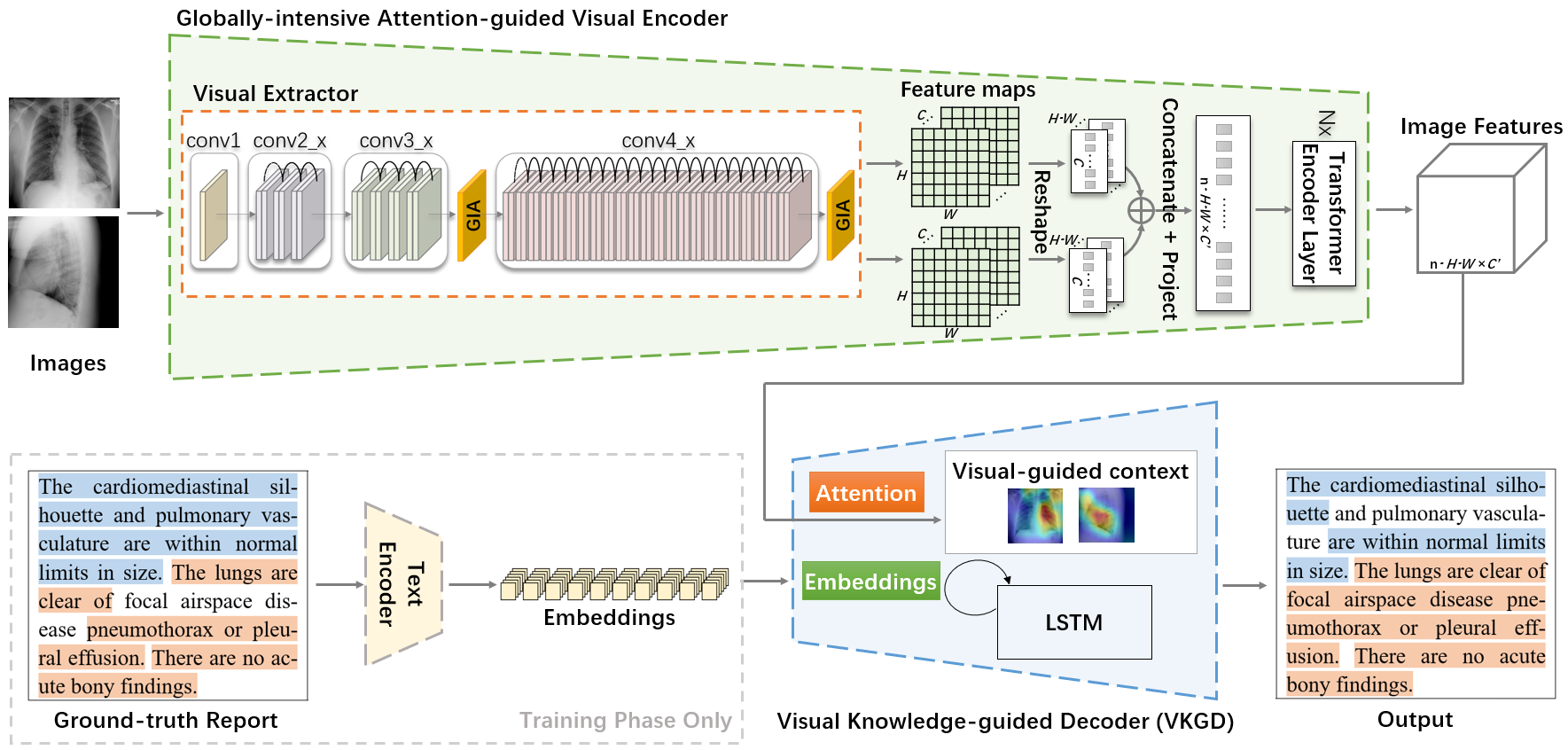}}
	\caption{The overall architecture of the proposed Intensive Vision-guided Network (IVGN). One or several radiology images are first visually encoded by the Globally-intensive Attention-guided Visual Encoder, which consists of a visual extractor based on a Globally-intensive Attention (GIA) module and several Transformer encoder layers. Then, the extracted image features are fed into the Visual Knowledge-guided Decoder (VKGD) for final report generation. In this decoder, the importance of image features for different regions and the importance of the previously predicted words are learned through an attention mechanism, resulting in a visual-guided context. Then, the visual-guided context, along with the text embeddings, are sent into a classical LSTM that focuses on mining associations between the previously predicted words and the current image. In the training phase, text embeddings come from the corresponding grouth-truth report of the input images, while in the inference phase, text embeddings refer to embeddings of the previously generated word. The decoder outputs words one by one, and finally forms a complete report. Best viewed in color.}
	\label{fig_4:Architecture}
\end{figure*}

The overall architecture of our proposed model Intensive Vision-guided Network (IVGN) is shown in Figure \ref{fig_4:Architecture}. Considering that some patients may need to have more than one radiology image taken during actual clinical examination, our model is designed to receive one or more radiology images as input. These radiology images have been converted from original grayscale images to three-channel images in RGB format by channel copying. Since the automatic generation of radiology reports is an image-to-text task, we follow the sequence-to-sequence paradigm \cite{C34}. For the image part, a Globally-intensive Attention-guided Visual Encoder is designed for image feature extraction (Figure \ref{fig_4:Architecture} upper). In this encoder, a visual extractor is designed based on a Globally-intensive Attention (GIA) mechanism to emphasize and extract important features in multiple perspectives, i.e. depth view, space view and pixel view (Figure \ref{fig_4:Architecture} upper left, the yellow GIA module, detailed in Figure \ref{fig_5:GIA}). The elements at each spatial position in the extracted feature maps form a vector along the channel dimension, so the extracted feature maps ($C \times H \times W $) will be transformed into $H\cdot W$ vectors, each of length $C$. If $n$ images are used to generate the same report, these images are sent into the visual extractor separately for feature extraction, and their feature maps are transformed into vectors separately and then concatenated to $n\cdot H \cdot W$ vectors. Then, after a projection operation that transforms the vector length to $C'$, these vectors forms the visual tokens in the input source sequence of Transformer. After that, a standard Transformer \cite{C35,C36} is used as an encoder to learn the associations of each visual token with the other tokens in the sequence and further extract semantic features (Figure \ref{fig_4:Architecture} upper right). Then, for the report generation part, we propose a Visual Knowledge-guided Decoder (VKGD) (Figure \ref{fig_4:Architecture} lower middle), which focuses on learning the associations between previously generated words and the current image and adaptively uses relevant visual information to assist text generation. In the training phase, VKGD receives the encoding results from the above visual encoder, as well as the corresponding ground-truth report of the input images (Figure \ref{fig_4:Architecture} lower left) as input, adopting the Teacher-Forcing training mode \cite{C37,C38}. In the inference phase, instead of taking the embeddings of the ground-truth report as input, VKGD takes the embeddings of the previously generated word as the text embeddings. VKGD uses an attention mechanism and a classical LSTM to iteratively generate visual-guided context, and finally forms the report. The Beam Search decoding strategy is used for decoding the generated reports during the inference phase. It selects multiple (rather than one) outputs with the maximum conditional probability and adds them to the candidate output sequence for the next time step. By doing this, the search space of candidate outputs for the generated text is enlarged, which is more conducive to generating more accurate reports. The details of the encoder and decoder are described in the following subsections.

\subsection{Globally-intensive Attention-guided Visual Encoder}

As shown in the upper part of Figure \ref{fig_4:Architecture}, one or several radiology images are first visually encoded by the Globally-intensive Attention-guided Visual Encoder which consists of a Globally-intensive Attention-based visual extractor and a standard Transformer.

\subsubsection{Globally-intensive Attention-based visual extractor}

The visual extractor takes ResNet-101 \cite{C39} as the backbone network. As mentioned above, focusing only on single-view modeling of radiology images while ignoring the information hidden in other perspectives may lead to an incomplete understanding of the semantics of the image. A human expert will examine the radiology images from multiple views to reconstruct the multi-view structure of organs. In order to model such multi-view examination into the network, we design a Globally-intensive Attention (GIA) module, which is embedded behind the last convolution layer of $conv3\_x$ and $conv4\_x$ of ResNet-101, as shown in Figure \ref{fig_4:Architecture}.

\begin{figure}[!t]
	\centerline{\includegraphics[width=0.9\columnwidth]{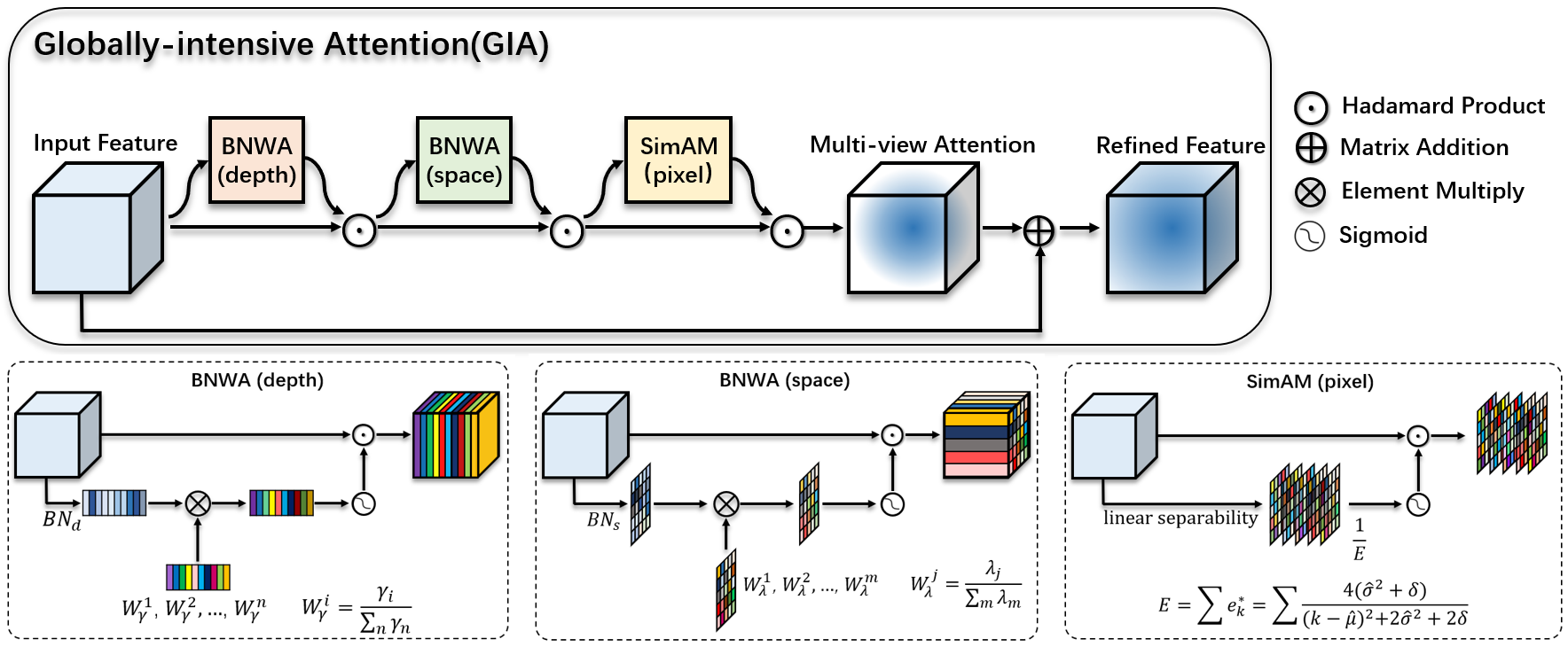}}
	\caption{The upper part shows the structure of the GIA module. The features extracted from the last convolution layer of $conv3\_x$ or $conv4\_x$ of ResNet-101 are first passed through a depth-view Batch Normalization-guided Weight Adapter (BNWA) submodule, and then a space-view BNWA submodule, to estimate the importance of depth weight and space weight, respectively. Then, after weighting the learned importance, the new features are then passed through a pixel-view SimAM submodule for importance learning of pixels in the whole feature maps. In this way, the module acquires multi-view attention. Finally, we use a residual structure to alleviate the problem of vanishing gradients \protect\cite{C39}. The bottom half of the figure are the schematic graphs of the depth-wise BNWA, space-view BNWA and pixel-view SimAM, respectively. The implementation details of these three submodules are described in Section 3.2.1. Best viewed in color.}
	\label{fig_5:GIA}
\end{figure}

The structure of the GIA module are shown in Figure \ref{fig_5:GIA}. To stimulate salient features and suppress indistinctive features, inspired by \citename{C40}\citeyear{C40}, we apply a Batch Normalization-guided Weight Adapter (BNWA) to scale the depth and space weights of the model. We hope that the model can learn more salient features of the radiology images from both depth view and space view, so we integrate both depth-view BNWA and space-view BNWA into our model. Experimental results show that the effect of using both views' BNMA  is indeed better than that of using only either view of them,  as described in the Results and Analysis section.

In addition, unlike natural images, where pixels can be naturally distinguished from each other, radiology images are grayscale images. Therefore, pixel-wise attention cannot be calculated directly based on the indistinguishable pixel values. \citename{C41}\citeyear{C41} propose that more important neurons have more obvious spatial suppression \cite{C42}, which can be measured by whether the neurons have greater linear separability with the other neurons. Inspired by it, we compute the linear separability of each pixel with the surrounding pixels in the feature maps to measure the importance of each pixel. For the first time, we apply linear separability measures to pixels in the feature maps of radiology images. We explore the activity of each pixel to enhance pixel-view attention.

Since depth-view BNWA, space-view BNWA and pixel-view attention enhance feature representation from different perspectives, our GIA module incorporates all three submodules. Our intuition is that serial stacking of the submodules will lead to a deeper network structure than parallelism, and thus is more conducive to the extraction of multi-level semantic information. Therefore, we stack the three submodules in series. Since the GIA module is connected after the $Conv3\_x$ and $Conv4\_x$ layers of ResNet-101, and the previous deep network layers has already carried out specific spatial feature extraction, we believe that salient feature extraction on depth-view is more urgent than that on space-view at this time. Therefore, we place the depth-view BNWA attention learning at the forefront of the GIA module, followed closely by the space-view BNWA. Compared with the above two submodules, pixel-view SimAM extracts more comprehensive attention information. Therefore, we place it at the end of the GIA module, allowing the model to comprehensively learn pixel-view attention information after processing the depth-view and space-view information, as shown in Figure \ref{fig_5:GIA}. Experiments have proved that the effect of serializing these three submodules is far better than that of paralleling them, and the effect of serializing all three is better than that of only taking two or even one of them. In addition, we verify through experiments that the sequence of these three submodules has a certain impact on the performance of salient feature extraction, and the sequence of depth-space-pixel is optimal overall, as detailed in the Results and Analysis Section.

The specific process of the GIA module is as follows. First of all, a scaling factor from batch normalization (BN) \cite{C43}, as shown in Equation (\ref{e2}), is used to measure the variance of channels and indicate their importance.

\begin{equation}
	B_{out}=BN(B_{in})=\gamma\frac{B_{in}-\mu_{\mathcal{B}}}{\sqrt{\sigma_{\mathcal{B}}^{2}+\epsilon}}+\beta \,, \label{e2}
\end{equation}

\noindent where $B_{in}$ represents the input value before $BN$ operation over a mini batch $\mathcal{B}$, $B_{out}$ represents the value after $BN$ operation, $\mu_{\mathcal{B}}$ and $\sigma_{\mathcal{B}}$ are the mean and standard deviation of $B_{in}$ in the mini batch $\mathcal{B}$, respectively, and $\gamma$ and $\beta$ are trainable scaling and shifting parameters.

Specifically, suppose that we obtain the image features $F$ of a radiology image $I$ after passing it through the $conv1$, $conv2\_x$, and $conv3\_x$ layers of ResNet-101, then the depth-view BNWA can be defined by

\begin{equation}
	F_d=sigmoid(W_\gamma(BN_d(F)))\odot F \,, \label{e3}
\end{equation}

\noindent where $F_d$ is the output features of depth-view BNWA, $W_{\gamma}=\{W_{\gamma}^{1}, W_{\gamma}^{2}, ..., W_{\gamma}^{n}\}$, where $W_{\gamma}^{i}=\gamma_{i}/\sum_{n}\gamma_{n}$ is the channel weight, where  $\gamma_{i}$ represents the trainable scaling factor of the $i$th channel, $BN_d(\cdot)$ is the batch normalization operated on depth/channel dimension, and $\odot$ is the hadamard product.

The output features of depth-view BNWA submodule are then sent to space-view BNWA submodule. Similarly, output features $F_s$ can be obtained by the following space-view BNWA operation

\begin{equation}
	F_s=sigmoid(W_\lambda(BN_s(F_d)))\odot F_d \,,
\end{equation}

\noindent where $W_{\lambda}=\{W_{\lambda}^{1}, W_{\lambda}^{2}, ..., W_{\lambda}^{m}\}$, where $W_{\lambda}^{j}=\lambda_{j}/\sum_{m}\lambda_{m}$ is the space weight, and $\lambda_{j}$ represents the trainable scaling factor of the $j$th pixel in the space. $BN_s(\cdot)$ is the batch normalization operated on space dimension, which specifically means that all pixels on each feature map are flattened into a one-dimensional vector for batch normalization operation, and $\odot$ is the hadamard product.

As mentioned above, features after weight scaling will enter a SimAM submodule to learn the attention weight of pixels, which can be described as

\begin{equation}
	F_{sim} =sigmoid(\frac{1}{E})\odot F_{s} \,,
\end{equation}

\noindent where $E$ groups all $e_{k}^{*}$ across channel and spatial dimensions, and $e_{k}^{*}$ is the minimal energy of the $k$th pixel, which is defined as

\begin{equation}
	e_{k}^{*}=\frac{4(\hat{\sigma}^{2}+\delta)}{(k-\hat{\mu})^{2}+2\hat{\sigma}^{2}+2\delta} \,,
\end{equation}

\noindent where  $\hat{\mu}=\frac{1}{M}{\textstyle\sum_{i=1}^{M}x_{i}}$ and $\hat{\sigma}^{2}=\frac{1}{M}{\textstyle\sum_{i=1}^{M}(x_{i}-\hat{\mu}})^{2}$ are mean and variance of all pixel values of the feature map in each channel, respectively, $x_{i}$ is the $i$th pixel value of the feature map, $M$ is the number of pixels of the feature map in each channel, and $\delta$ is the regularization parameter \cite{C41}.

After that, a residual connection will be made between the original input features and $F_{sim}$, and finally the output feature $F_{GIA}$ of the GIA module will be obtained. The pseudocode of the GIA module is shown in Algorithm \ref{alg:1}.

The resulting $F_{GIA}$ will then be fed into the $conv4\_x$ layer of ResNet-101. Similarly, at the end of the $conv4\_x$ layer, the feature maps will pass through another GIA module to further enhance multiple perspective attention.

\begin{algorithm}[!t]
	
	\caption{The pseudocode of the GIA module (pytorch-like implementation). }
	\label{alg:1}
	\footnotesize{\textbf{\texttt{
				\\
				\textcolor[RGB]{34,139,34}{
					""" \\
					Input:\\
					F -- image features after passing through the conv1, conv2\_x, and conv3\_x layers of ResNet-101, [bs, c, h, w]; \\
					delta -- coefficient $\delta$ in Equation (6)\\
					\\
					Output: F\_{GIA}\\
					""" \\
				}
	}}}
	\textbf{\texttt{
			\begin{algorithmic}\\
				bs, c, h, w = feats.shape \\
				residual = feats \\
				\\
				\textcolor[RGB]{34,139,34}{\# depth-view BNWA} \\
				att = batch\_norm(feats) \\
				gamma = \textcolor[RGB]{238,180,34}{abs}(batch\_norm.weight) / \textcolor[RGB]{238,180,34}{sum}(\textcolor[RGB]{238,180,34}{abs}(batch\_norm.weight)) \\
				att = att.permute(0, 2, 3, 1) \\
				att = (gamma * att).permute(0, 3, 1, 2) \\
				feats = feats * sigmoid(att) \\
				\\
				\textcolor[RGB]{34,139,34}{\# space-view BNWA} \\
				att = batch\_norm(feats.reshape(bs, c, -1).permute(0, 2, 1)) \\
				lambda = \textcolor[RGB]{238,180,34}{abs}(batch\_norm.weight) / \textcolor[RGB]{238,180,34}{sum}(\textcolor[RGB]{238,180,34}{abs}(batch\_norm.weight)).permute(0, 2, 1)\\
				att = (lambda * att).view(bs, c, h, w) \\
				feats = feats * sigmoid(att) \\
				\\
				\textcolor[RGB]{34,139,34}{\# pixel-view SimAM} \\
				sigma\^{}2 = (x - x.mean(\textcolor{cyan}{dim} = [2, 3])) ** 2 \\
				E\_inv = sigma\^{}2 / (4 * (sigma\^{}2.sum(\textcolor{cyan}{dim} = [2, 3]) / (h * w - 1) + delta)) + 0.5 \\
				\\
				feats = feats * sigmoid(E\_inv) \\
				F\_{GIA} = residual + feats \\
				\\
				\textcolor[RGB]{255,105,180}{return} F\_{GIA} \\
			\end{algorithmic}
	}}
\end{algorithm}

\subsubsection{Transformer}

 Transformer is a deep learning model based on the self-attention mechanism, which is more effective than the traditional CNN model in mining and processing long-distance dependencies, more sensitive to the location information of objects, and more conducive to global modeling. With the above CNN (ResNet-101) and GIA-based visual extractor, the model can learn important local features in the image. However, the features of lesions such as cardiomegaly and atelectasis in the X-ray images of this task cover a wide range of locations, and feature extraction using CNN alone may not be able to effectively capture the global information of these lesions. Therefore, we introduce a Transformer encoder behind the above CNN and GIA-based visual extractor to help capture image features on a larger scale to better understand global semantic information.

Specifically, after feature extraction by the above Globally-intensive Attention-based visual extractor, the obtained feature maps $X$ are transformed into an input source sequence to be fed into the subsequent Transformer. Specifically, the elements at each spatial position in the extracted feature maps form a vector along the channel dimension, so the extracted feature maps ($C \times H \times W $) will be transformed into $H\cdot W$ vectors, each of length $C$. If $n$ images are used to generate the same report, these images are sent into the visual extractor separately for feature extraction, and their feature maps are transformed into vectors separately and then concatenated to $n\cdot H \cdot W$ vectors. Then, after a projection operation that transforms the vector length to $C'$, these vectors form the the input source sequence $\{X_1,X_2,...,X_S\}$, where $S=n\cdot H\cdot W$. Subsequently, a standard Transformer encoder is used to learn the associations of each visual token $X_i$ with the other tokens in the sequence and extract deeper semantic features. Specifically, the process can be formulated as

\begin{equation}
	Y=\{Y_1,Y_2,...,Y_S\}=F_t(X_1,X_2,...,X_S) \,,
\end{equation}

\noindent where $F_t(\cdot)$ represents the Transformer encoder, and $Y=\{Y_1,Y_2,...,Y_S\}$ are the final output features of the encoder.

\subsection{Visual Knowledge-guided Decoder}

In clinical medicine, each physiological part of the human organ will have its unique medical characteristics. Therefore, in radiology reports, features are always closely related to their corresponding physiological parts. So in many cases, it is expected that there is a special correspondence between adjacent words in radiology reports. For example, if the former word is a term describing a certain physiological part, then the subsequent word may correspond to a certain feature of that physiological part. In this case, the adjacent words can be predicted reliably from the same guided vision information. On the other hand, the text generator likely requires little or no visual information from the image to predict non-visual words such as ``a'', ``and'' and other conjunctions, etc. In this situation, the text generator should not rely too much on visual information to generate the text. Hence, to better match image features with the corresponding physiological parts, we propose a Visual Knowledge-guided Decoder (VKGD), which focuses on mining associations between the previously predicted words and the corresponding region of the current image.

As shown in Figure \ref{fig_4:Architecture}, in the training phase, the ground-truth radiology report of the input images will first go through a typical Text Encoder to be converted into the corresponding text embeddings. In the inference phase, the text embedding comes from the embeddings of the previously generated word. These text embeddings are then fed into VKGD as input along with the output of the Globally-intensive Attention-guided Visual Encoder.

\begin{figure}[!t]
	\centerline{\includegraphics[width=1\columnwidth]{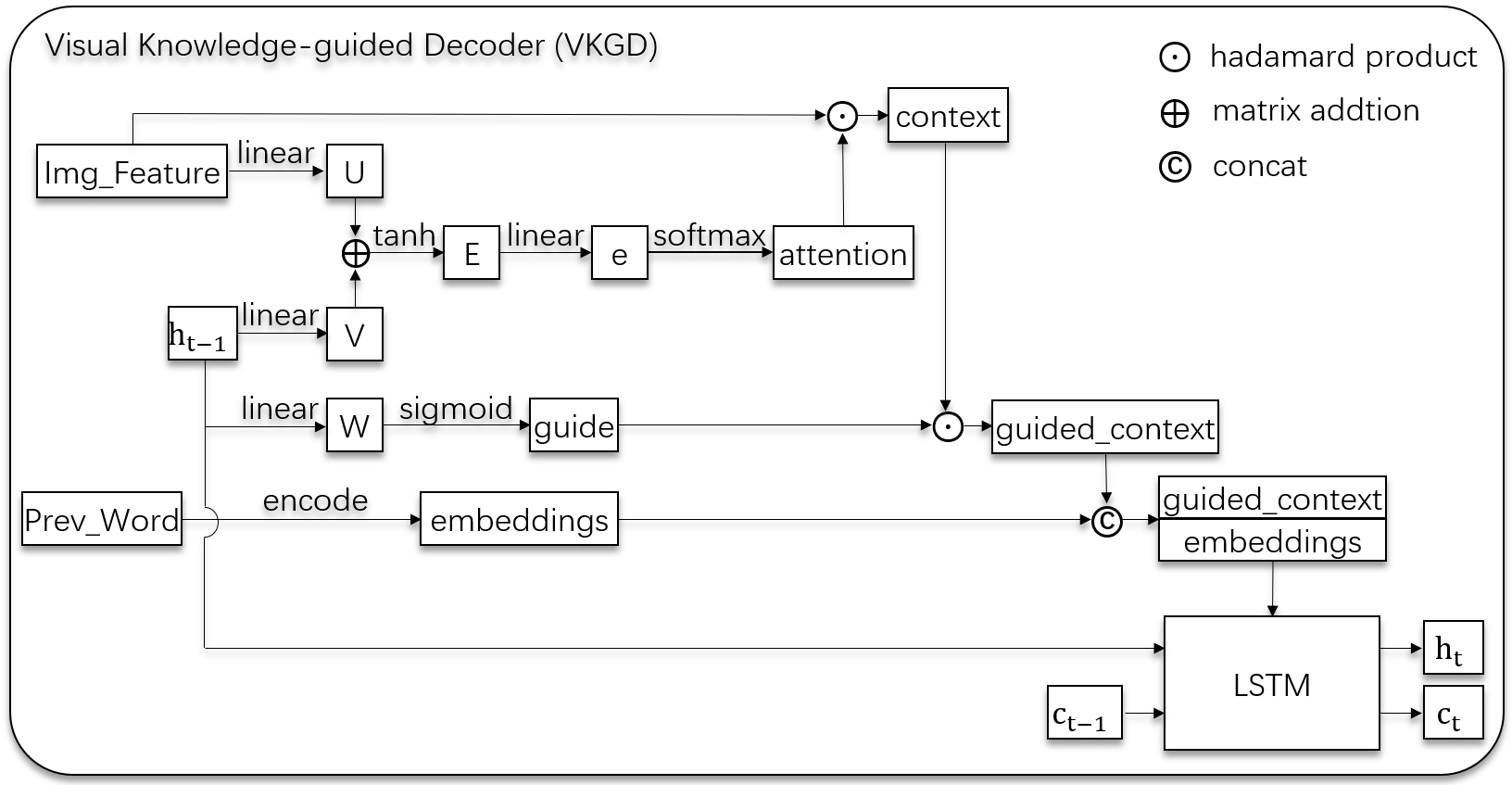}}
	\caption{The implementation detail of our proposed Visual Knowledge-guided Decoder (VKGD). }
	\label{fig_6:VKGD}
\end{figure}

The implementation details of VKGD are shown in Figure \ref{fig_6:VKGD}. VKGD is mainly composed of a LSTM network \cite{C44} and an attention mechanism. In the LSTM, a new word is generated at each step based on the context vector, the previous hidden state and the word generated at the previous step, and finally the whole report is generated. The specific implementation of LSTM follows the study of \citename{C45}\citeyear{C45}, which will not be described here. The reason of using LSTM network instead of Transformer as the decoder is that radiology reports are generally long in length, and Transformer has restrictions on the length of input. Too long text sequences will lead to daunting time and memory costs, while LSTM has no such restrictions. In addition, Transformer has a large amount of parameters, while LSTM is lightweight. We also conducted a comparative experiment on these two decoders. Experimental results show that the effect of using LSTM as the decoder is far better than that of using Transformer, detailed in the Results and Analysis section.

Besides LSTM, an attention module is designed to measure the contextual association between the previous output word and the features of the current input image. Specifically, as shown in Figure \ref{fig_6:VKGD}, the image features $Img\_Feature$ extracted by the previous Globally-intensive Attention-guided Visual Encoder and the previous hidden state $h_{t-1}$ of the LSTM are separately transformed by a linear transformation. Then, these two transformed features are added together and sent into a linear layer to learn the correlation between the image features at different positions and the previously predicted words, so as to obtain the relative importance of each position in the image. Then, the $context$ information $C$ can be obtained by hadamard multiplying the image feature at each position and its corresponding attention information, as described in

\begin{equation}
	C =Y \odot A \,, \label{e8}
\end{equation}

\noindent where $Y=\{Y_1,Y_2,...,Y_S\}$ are the features of the $S$ image positions extracted by the Globally-intensive Attention-guided Visual Encoder, $\odot$ is the hadamard product, and $A=\{A_1,A_2,...,A_S\}$ are the attention information of the $S$ image positions. The attention information of the $i$th position $A_i$ can be calculated by

\begin{equation}
	A_i =\frac{exp(F_{a}(Y_{i},h_{t-1}))}{{\textstyle\sum_{k=1}^{S}exp(F_{a}(Y_{k},h_{t-1}))}} \,, \label{e9}
\end{equation}

\noindent where $h_{t-1}$ is the previous hidden state of the LSTM, and $F_{a}(\cdot)$ represents the linear layer.

To emphasize the impact of the previous hidden state $h_{t-1}$ on the generation of new word, we further incorporate it into the final context information $guided\_context$ $g_c$, which is then fed into the LSTM. $g_c$ can be calculated by

\begin{equation}
	g_c =sigmoid(F_g(h_{t-1})) \odot C \,, \label{e10}
\end{equation}

\noindent where $F_g(\cdot)$ represents a linear layer, and $\odot$ is the hadamard product.

Finally, we use a deep output layer to calculate the output probability of each word given the LSTM state, the context vector and the previous word. Suppose that the text sequence of the target report is $R=\{r_1,r_2,...,r_T\}$, then given a radiology image $I$, the probability of final output $R$ is

\begin{equation}
	p(R|I)=\prod_{t=1}^{T}p(r_{t}|r_{1},...,r_{t-1},I) \,,
\end{equation}

\noindent where, at time $t$, for the given LSTM state $h_{t-1}$, context vector $g_c$ and the previous word $r_{t-1}$, the conditional probability of the current output of the word $r_t$ can be formulated as

\begin{equation}
	p(r_{t}|r_{1},...,r_{t-1},I) \propto exp(L_{o}(Er_{t-1}+L_{h}h_{t-1}+L_{c}g_c)) \,,
\end{equation}

\noindent where $L_o$, $E$, $L_h$ and $L_c$ are learned parameters initialized randomly.

The model is trained to minimize $-p(R|I)$ through the negative conditional log likelihood of $R$ given the image $I$, as described in

\begin{equation}
	\theta^{*}=\underset{\theta}{arg min}\sum_{t=1}^{T}-log\, p(r_{t}|r_{1},...,r_{t-1},I;\theta) \,,
\end{equation}

\noindent where $\theta$ represents the model parameters.

\section{Experiments}

\subsection{Datasets}
We performed extensive experiments on two commonly-used medical image report datasets, \textsc{IU X-Ray} \cite{C47} and MIMIC-CXR \cite{C48}, to verify the model's effectiveness. Both datasets are third-party publicly available database with preexisting institutional review board (IRB) approval.

\textbf{\textsc{IU X-Ray}} is a publicly available radiological dataset collected by Indiana University, with 7,470 frontal and lateral-view chest X-ray images and 3,955 reports. The reports include \emph{Findings}, \emph{Impression}, \emph{Tags}, \emph{Comparison} and \emph{Indication}. Following \citename{C7}\citeyear{C7}, we excluded images without reports and there are 5,910 images and 2,955 reports left for this study. Following \citename{C5}\citeyear{C5}, we split the data into training/validation/test set by 7:1:2 of the dataset, and took the \emph{Findings} sections as the target captions to be generated.

\textbf{MIMIC-CXR} is the largest radiology image dataset so far, sourcing from the Beth Israel Deaconess Medical Center between 2011-2016. We followed \citename{C8}\citeyear{C8} to adopt an alpha version of 473, 057 Chest X-ray images and 206, 563 reports from 63, 478 patients. We adopted the official split of training/validation/test set, and took the \emph{Findings} section as the target captions to be generated.

\subsection{Implementation details}

Our model can take one or more images of each patient as input. ResNet-101 pre-trained on ImageNet \cite{C49} is adopted as the backbone of the visual extractor to extract features from input images reshaped to 224 $\times$ 224. Feature maps obtained by the visual extractor are 7$\times$7 in spatial dimension, with 2,048 channels. Subsequently, elements at each of these 7$\times$7 spatial positions form a vector along the channel dimension, resulting a total of 49 vectors, each of length 2,048. If there are $n$ images to generate the same report, the transformed vectors of these images are concatenated to $n$$\times$49 vectors. The rest components of the model are trained from scratch. The hidden dimension and the number of heads for the Transformer Encoder are 512 and 8, respectively. We set the hidden dimension of the LSTM in VKGD to 512. We trained our model using cross-entropy loss with ADAM optimizer \cite{C50}. We adopted L2 regularization and set the weight decay to 5e-5. We set the initial learning rate for ResNet-101, the GIA module and the rest part of our model to 0.001, 5e-5 and 0.01 respectively on \textsc{IU X-Ray}, and 5e-5, 5e-5 and 1e-4 respectively on MIMIC-CXR. We reduced the learning rate by 0.8 every 10 epochs on \textsc{IU X-Ray} and every 20 epochs on MIMIC-CXR. We set batch size to 16 and trained 50 epochs on \textsc{IU X-Ray}, set batch size to 12 and trained 100 epochs on MIMIC-CXR, and finally selected the models with the best BLEU-4 on the validation set as the inference models. And we set beam size, which is the number of outputs with the highest conditional probability at each time step to be added to the candidate output sequence for the next time step in the Beam Search decoding strategy, to 3 at inference time. All these hyperparameters were optimized in the validation set. The specific optimization process and the impact of different values of each hyperparameter on model’s performance are detailed in Section S2 in the supplementary material.

\subsection{Baselines}

We used the following baseline: a ResNet-101 pre-trained on ImageNet as the visual encoder and a typical Transformer with 3 layers, 8 heads, and 512 hidden units as the decoder, which is randomly initialized. In addition, we compared our model with previous studies, including HRGR \cite{C7}, SentSAT+KG \cite{C12}, CoAtt \cite{C6}, PKERRG \cite{C13}, CMAS-RL \cite{C9}, R2Gen \cite{C5}, CMN \cite{C16}, KERP \cite{C10}, PPKED \cite{C14}, \textsc{$\mathcal{M}^2$Tr.Progressive} \cite{C15}, Multicriteria\cite{d3}, KM\cite{d2} and CCR \cite{C8}. We adopted the results reported in their papers for comparison.

\subsection{Evaluation metrics}

Following most of the studies on this task, we evaluated our proposed method, the baseline models and previous models using the conventional natural language generation (NLG) metrics, which include BLEU \cite{C51}, METEOR \cite{C52} and ROUGE-L \cite{C53}. Among them, BLEU is a metric that measures the accuracy of the generated text. Specifically, BLEU-N measures how many consecutive N words in the generated text appear in the ground-truth text. Thus, BLEU-1 measures the accuracy at the word level, BLEU-2 and BLEU-3 measure the accuracy at the phrases and expressions level, while the higher-order BLEU-4 measures the accuracy of long expressions and sentences, as well as the logical semantic coherence within a single long sentence. METEOR is an indicator that focuses on measuring the fluency of sentences. It converts the generated text and the ground-truth text into word sets respectively, and uses semantic resources such as WordNet to measure the similarity between the generated and ground-truth text by considering factors such as word-level semantic similarity, synonyms, inflection, and word sequence. In addition, it also takes into account factors such as misaligned words, phrase repetition, and length differences between the generated and ground-truth texts. ROUGE-L measures the similarity between the generated and ground-truth text by calculating the overlap rate of the longest common subsequence between the predicted and ground-truth text. It can capture the integrity of the overlapping parts in the generated and ground-truth text, and is also an indicator that focuses on measuring the fluency of sentences and the logical relationship between words and sentences.

It is worth mentioning that radiology reports differ greatly from other generated texts in terms of text requirements and evaluation. In addition to meeting the basic requirements of smoothness and fluency, radiology reports emphasize the professional accuracy of the text more than ordinary reports or text descriptions. That is, the medical terminology, disease terminology and specific physiological parts mentioned in the radiology report need to be very precise, and a slight deviation will greatly affect the diagnosis results. Therefore, following \citename{C5}\citeyear{C5} and \citename{C8}\citeyear{C8}, we adopted the clinical efficacy (CE) metrics, which include precision, recall and F1-score to further evaluate the quality of the generated reports. The calculation method for the precision, recall, and F1-score metrics in our report generation task is quite different from that of ordinary classification tasks. The specific calculation process is as follows: Firstly, using CheXpert labeler\footnote{https://github.com/MIT-LCP/mimic-cxr/tree/master/txt/chexpert}, the information about the 14 different categories related to thoracic diseases and support devices defined by CheXpert \cite{C54} is extracted from the ground-truth report and generated report. Specifically, if the report contains a diagnosis that the patient is suffering from some categories of the 14 diseases, the corresponding categories are considered positive. Conversely, if the report contains a diagnosis that the patient does not have some categories of the 14 diseases, the corresponding categories are considered negative, and the disease categories not mentioned are not considered. Then, using this positive and negative information, precision, recall and F1-score are calculated for each category, and the overall precision, recall and F1-score for a particular patient's report is the average of the precision, recall and F1-score over the 14 categories.

\section{Results and Analysis}

\subsection{Comparison with previous studies}

\begin{table*}
	\scriptsize
	\caption{Comparisons of our model with previous studies on the \textsc{IU X-Ray} and MIMIC-CXR test set with respect to language generation (NLG) and  clinical efficacy (CE) metrics. BL-n denotes BLEU score using up to n-grams; MTR and RG-L denote METEOR and ROUGE-L, respectively. P, R and F1 represent precision, recall and F1-score, respectively. IVGN is our proposed model. Best results are in bold.}
	\label{table:comparisons_with_previous}
	\begin{center}
		\begin{tabular}{c|c|cccccc|ccc}
			\hline
			\multirow{2}{*}{Dataset} & \multirow{2}{*}{Model} & \multicolumn{6}{c|}{NLG Metrics} & \multicolumn{3}{c}{CE Metrics}\\
			& & BL-1 & BL-2 & BL-3 & BL-4 & MTR & RG-L & P & R & F1 \\
			\hline
			\multirow{13}{*}{\textsc{IU X-Ray}} & HRGR & 0.438 & 0.298 & 0.208 & 0.151 & - & 0.322 & - & - & - \\
			& SentSAT+KG & 0.441 & 0.291 & 0.203 & 0.147 & - & 0.367 & - & - & - \\
			& CoAtt & 0.455 & 0.288 & 0.205 & 0.154 & - & 0.369 & - & - & - \\
			& PKERRG & 0.450 & 0.301 & 0.213 & 0.158 & - & 0.384 & - & - & - \\
			& CMAS-RL & 0.464 & 0.301 & 0.210 & 0.154 & - & 0.362 & - & - & - \\
			& R2Gen & 0.470 & 0.304 & 0.219 & 0.165 & 0.187 & 0.371 & - & - & - \\
			& CMN & 0.475 & 0.309 & 0.222 & 0.170 & 0.191 & 0.375 & - & - & - \\
			& KERP & 0.482 & 0.325 & 0.226 & 0.162 & - & 0.339 & - & - & - \\
			& PPKED & 0.483 & 0.315 & 0.224 & 0.168 & 0.190 & 0.376 & - & - & - \\
			& \textsc{$\mathcal{M}^2$Tr.Progressive} & 0.486 & 0.317 & 0.232 & 0.173 & 0.192 & 0.390 & - & - & - \\
			& Multicriteria & 0.496 & 0.319 & 0.241 & 0.175 & - & 0.377 & - & - & - \\
			& KM & 0.496 & 0.327 & 0.238 & 0.178 & - & 0.381 & - & - & - \\
			& IVGN &  \textbf{0.523} & \textbf{0.357} & \textbf{0.258} & \textbf{0.191} & \textbf{0.226} & \textbf{0.405} & - & - & - \\
			\hline
			\multirow{8}{*}{MIMIC-CXR} & CCR & 0.313 & 0.206 & 0.146 & 0.103 & - & \textbf{0.306} & - & - & - \\
			& Multicriteria & 0.351 & 0.223 & 0.157 & \textbf{0.118} & - & 0.287 & - & - & - \\
			& R2Gen & 0.353 & 0.218 & 0.145 & 0.103 & 0.142 & 0.277 & 0.333 & 0.273 & 0.276 \\
			& CMN & 0.353 & 0.218 & 0.148 & 0.106 & 0.142 & 0.278 & 0.334 & 0.275 & 0.278 \\
			& PPKED & 0.360 & 0.224 & 0.149 & 0.106 & \textbf{0.149} & 0.284 & - & - & - \\
			& KM & 0.363 & 0.228 & 0.156 & 0.115 & - & 0.284 & \textbf{0.458} & 0.348 & 0.371 \\
			& \textsc{$\mathcal{M}^2$Tr.Progressive} & \textbf{0.378} & 0.232 & 0.154 & 0.107 & 0.145 & 0.272 & 0.240 & 0.428 & 0.308 \\
			& IVGN & 0.377 & \textbf{0.236} & \textbf{0.158} & 0.112 & 0.144 & 0.280 & 0.415 & \textbf{0.442} & \textbf{0.398} \\
			\hline
		\end{tabular}
	\end{center}
	
\end{table*}

The comparison results of our model (IVGN) with models in previous studies are shown in Table \ref{table:comparisons_with_previous}. On \textsc{IU X-Ray}, our method significantly outperforms methods in previous studies in all metrics. This shows that our model performs better not only in terms of word and phrase generation, but also in terms of long sentences and the logic between sentences. On MIMIC-CXR, our method surpasses the existing methods in BL-2 and BL-3, and achieves comparable performance to the state-of-the-art methods \textsc{$\mathcal{M}^2$Tr.Progressive} \cite{C15}, KM \cite{d2} and PPKED \cite{C14} in other NLG metrics. The reason why the MTR and RG-L metrics of the model are not optimal may be that the order of lesions or sentences in the reports generated by our model is not strictly consistent with the order in the ground-truth reports although they can accurately describe the situation of patients' radiology images (to be discussed in detail in Section 5.3.3). It is worth mentioning that on MIMIC-CXR, our method greatly exceeds the previous methods in three CE metrics, namely precision, recall and F1-score (an improvement of 17.5\%, 1.4\% and 9\% respectively compared with \textsc{$\mathcal{M}^2$Tr.Progressive}, an improvement of 8.1\%, 16.7\% and 12\% respectively compared with CMN, and an improvement of 8.2\%, 16.9\% and 12.2\% respectively compared with R2Gen). Our method has lower precision than that of KM, but it exceeds KM in the more comprehensive F1-score metric. CE metrics are specifically proposed for medical image reports generation \cite{C5,C54,C16}, which can effectively measure the clinical efficacy of medical image reports. In the calculation of CE metrics, words of 14 different kinds of diseases, including cardiomegaly, lung lesion, lung opacity, edema, pneumonia, atelectasis, pneumothorax, pleural effusion, and fracture, are extracted from the ground-truth report and the generated report, and then precision, recall and F1 are calculated respectively according to the extracted words. As shown in Table \ref{table:comparisons_with_previous}, our model achieves much higher precision and recall, which indicates that our model predicts much fewer false positive and false negative diseases, respectively. There are two main reasons for this. First, our model conducts multi-view (depth-view, space-view and pixel-view) analysis on images, which to some extent simulates the multi-view analysis and remodeling of human organs and lesions by clinicians when examining radiology images. Therefore, the model predicts the location of lesions and their corresponding physiological parts more accurately. Second, the model can guide the generation of next word based on the previously predicted word and visual features, which may be the physiological parts detected by the visual extractor. So there is a good correlation between the specific physiological parts and their corresponding lesions in the report. Both reasons make our model have obvious advantages in the three metrics of clinical efficacy, and the generated reports are more reliable. On the other hand, we noticed that the precision, recall and F1-score of all models are relatively low compared to that of ordinary classification tasks. Therefore, we conducted further analysis in Section 5.3.4.

\subsection{Ablation study: the effect of the components and submodules}

To evaluate the contributions of the two components (i.e. the GIA module and VKGD) and the submodules in both components in our proposed model (IVGN), we conducted extensive experiments on \textsc{IU X-Ray} and the results are shown in Table \ref{table:ablation_study}, Table \ref{table:ablation_study_GIA} and Table \ref{table:ablation_study_VKGD}.
	
In Table \ref{table:ablation_study}, We compared our proposed method with the baseline method and integrated the GIA module and VKGD separately for comparison. The baseline model includes a ResNet-101 pre-trained on ImageNet as the visual encoder and a typical Transformer with 3 layers, 8 heads, and 512 hidden units as the decoder, which is randomly initialized. Model-1 adds the GIA module to the visual extractor of the baseline model. The GIA module adopts the structure shown in Figure \ref{fig_5:GIA}, where three submodules are processed serially in the sequence of depth-view / space-view / pixel-view. Model-2 replaces the typical Transformer in the baseline model with VKGD. And our IVGN includes both the GIA module and VKGD.

In addition, to further explore the impact of submodule combination, serialization, and parallelism of the three submodules (depth-view BNWA, space-view BNWA and pixel-view SimAM) in the GIA module on the overall performance of the model, we conducted a more comprehensive experimental verification on all the possible cases of submodule combination, serialization, and parallelism, including their logical sequence, as shown in Table \ref{table:ablation_study_GIA}. For VKGD, we also verified through experiment the important contribution of the attention mechanism in it, as shown in Table \ref{table:ablation_study_VKGD}.

\begin{table*}
	\caption{Performance of the baseline model and models with different components in ablation study on the \textsc{IU X-Ray} test set with respect to language generation (NLG) metrics. BL-n denotes BLEU score using up to n-grams; MTR and RG-L denote METEOR and ROUGE-L, respectively. ``GIA'' represents the GIA module adopting the structure shown in Figure \ref{fig_5:GIA}, where three submodules are processed serially in the sequence of depth-view / space-view / pixel-view. ``VKGD'' is the proposed Decoder for reports generation. IVGN is our proposed model. Best results are in bold.}
	\label{table:ablation_study}
	\begin{center}
		\begin{tabular}{c|c|c|cccccc}
			\hline
			\multirow{2}{*}{Model} & \multirow{2}{*}{GIA} & \multirow{2}{*}{VKGD} & \multicolumn{6}{c}{NLG Metrics} \\
			& & & BL-1 & BL-2 & BL-3 & BL-4 & MTR & RG-L  \\
			\hline
			Baseline & & & 0.396 & 0.254 & 0.179 & 0.135 & 0.164 & 0.342 \\
			Model-1 & \checkmark & & 0.456 & 0.297 & 0.215 & 0.166 & 0.190 & 0.359 \\
			Model-2 & & \checkmark & 0.469 & 0.322 & 0.242 & 0.188 & 0.179 & 0.377 \\			
			IVGN & \checkmark & \checkmark  & \textbf{0.523} & \textbf{0.357} & \textbf{0.258} & \textbf{0.191} & \textbf{0.226} & \textbf{0.405} \\
			
			\hline
			
		\end{tabular}
	\end{center}
\end{table*}

\subsubsection{The effect of the GIA module}

\begin{table*}
	\caption{Performance of models with different submodule combination, serialization, and parallelism schemes. Baseline* is Model-2 in Table \ref{table:ablation_study}, whose encoder is a ResNet-101 without adding any depth-view BNWA, space-view BNWA or pixel-view SimAM submodule. ‘d’, ‘s’, ‘p’ stands for the submodule of depth-view BNWA, space-view BNWA and pixel-view SimAM, respectively. For the ``parallel'' mode, the square bracket indicates that the enclosed two or three submodules are in parallel. For the ``serial'' mode, the execution sequence between submodules is the sequence in which their representative letters (‘d’, ‘s’, ‘p’) appear. The ``mix'' mode represents the cases where parallel and serial are mixed together. For example, ‘[ds]p’ means that depth-view BNWA and space-view BNWA are first operated in parallel and then connected in series with pixel-view SimAM. Note that in all these models, VKGD is used as the decoder. All models are trained and tested with the same hyperparameters, and no hyperparameter tuning has been performed. Best results are in bold.}
	\label{table:ablation_study_GIA}
	\begin{center}
		\begin{tabular}{c|c|r@{ : }c|cccccc}
			\hline
			Number of  & \multirow{2}{*}{Mode} & \multicolumn{2}{c|}{\multirow{2}{*}{Scheme}} & \multicolumn{6}{c}{NLG Metrics} \\
			submodules & & \multicolumn{2}{c|}{} & BL-1 & BL-2 & BL-3 & BL-4 & MTR & RG-L  \\
			\hline
			\multicolumn{4}{c|}{Baseline*} & 0.469 & 0.322 & 0.242 & 0.188 & 0.179 & 0.377 \\
			\hline
			\multirow{3}{*}{one submodule} & \multirow{3}{*}{-} & \#1 & d & 0.485 & 0.328 & 0.241 & 0.184 & 0.193 & 0.383 \\
			& & \#2 & s & 0.485 & 0.323 & 0.234 & 0.179 & 0.195 & 0.379 \\
			& & \#3 & p & 0.482 & 0.324 & 0.238 & 0.183 & 0.192 & 0.382 \\
			\hline
			\multirow{9}{*}{two submodules}& \multirow{3}{*}{parallel} & \#4 & [ds] & 0.493 & 0.330 & 0.241 & 0.185 & 0.196 & 0.383 \\
			& & \#5 & [dp] & 0.494 & 0.335 & 0.246 & 0.189 & 0.200 & 0.383 \\
			& & \#6 & [sp] & 0.494 & 0.329 & 0.238 & 0.181 & 0.200 & 0.379 \\
			\cline{2-10}
			& \multirow{6}{*}{serial} & \#7 & ds & 0.498 & 0.333 & 0.246 & 0.191 & 0.203 & 0.381 \\
			& & \#8 & sd & 0.489 & 0.333 & 0.246 & 0.189 & 0.195 & 0.385 \\
			& & \#9 & dp & 0.499 & 0.335 & 0.246 & 0.189 & 0.202 & 0.383 \\
			& & \#10 & pd & 0.491 & 0.326 & 0.241 & 0.187 & 0.201 & 0.383 \\
			& & \#11 & sp & 0.487 & 0.326 & 0.237 & 0.181 & 0.196 & 0.386 \\
			& & \#12 & ps & 0.483 & 0.326 & 0.239 & 0.184 & 0.193 & 0.381 \\
			\hline
			\multirow{13}{*}{three submodules}& parallel & \#13 & [dsp] & 0.495 & 0.332 & 0.244 & 0.188 & 0.200 & 0.387 \\
			\cline{2-10}
			& \multirow{6}{*}{mix} & \#14 & [ds]p & 0.492 & 0.321 & 0.230 & 0.174 & 0.193 & 0.377 \\
			& & \#15 & p[ds] & 0.499 & 0.345 & 0.254 & 0.191 & \textbf{0.219} & 0.393 \\
			& & \#16 & [dp]s & 0.495 & 0.343 & 0.251 & 0.189 & \textbf{0.219} & 0.395 \\
			& & \#17 & s[dp] & 0.493 & 0.329 & 0.244 & 0.191 & 0.201 & 0.382 \\
			& & \#18 & [sp]d & 0.505 & 0.346 & 0.253 & 0.190 & 0.214 & 0.393 \\
			& & \#19 & d[sp] & 0.492 & 0.329 & 0.240 & 0.183 & 0.196 & 0.381 \\
			\cline{2-10}
			& \multirow{6}{*}{serial} & \#20 & sdp & 0.508 & 0.342 & 0.251 & 0.193 & 0.205 & 0.394 \\				
			& & \#21 & spd & 0.508 & 0.343 & 0.248 & 0.188 & 0.211 & 0.389 \\
			& & \#22 & pds & 0.501 & 0.343 & 0.249 & 0.187 & 0.213 & 0.396 \\
			& & \#23 & psd & 0.507 & 0.342 & 0.250 & 0.190 & 0.207 & 0.394 \\
			& & \#24 & dps & 0.510 & 0.343 & 0.250 & 0.189 & 0.209 & 0.393 \\
			& & \#25 & \textbf{dsp} & \textbf{0.513} & \textbf{0.349} & \textbf{0.257} & \textbf{0.196} & 0.211 & \textbf{0.406} \\
			\hline
			
		\end{tabular}
	\end{center}
\end{table*}

Comparing Model-1 and the baseline model in Table \ref{table:ablation_study}, it can be clearly seen that the whole GIA module has brought significant contributions to the improvement of model performance in all metrics (up to 6\% in the BL-1 metric). This indicates that the whole GIA module is effective in learning words and phrases as well as long sentences and logical relationships between sentences.

The contributions of each submodule in the GIA module are shown in Table \ref{table:ablation_study_GIA}. It can be observed that, first of all, depth-view BNWA, space-view BNWA and pixel-view SimAM are all valid when participating in the feature extractor as separate submodules (Scheme \#1, \#2, \#3 vs. Baseline*, where Baseline* is Model-2 in Table \ref{table:ablation_study}). All three submodules contribute almost equally to the performance improvement, with space-view BNWA a bit weaker overall. The possible reason is that depth-view BNWA performs attention learning of channel weights, so it pays more attention to the relevance in the depth direction, which makes up for the weakness of 2D convolution learning in feature extraction in the depth direction. Therefore, after the addition of depth-view BNWA module, the learning and representation ability of the model in the depth direction has been improved. Similarly, pixel-view SimAM learns the importance of every pixel at the pixel-wise level, which also enables the model to learn the high-dimensional features of length, width and height in a more comprehensive way.

By comparing Scheme \#4 to \#12 with Scheme \#1 to \#3, and Scheme \#13 to \#25 with Scheme \#4 to \#12, it can be seen that the overall performances of three submodules integrated are better than those of two submodules, and the overall performances of two submodules integrated are better than those of a single submodule, regardless of whether these submodules are in serial, in parallel, or mixed. This means that the integration of more submodules into the visual extractor has a positive impact on model performance, for both the generation of words and phrases, and the feature extraction of long sentences and logical relations. This is exactly in line with our motivation to propose the model - We believe that comprehensive and multi-view (depth-view, space-view, pixel-view) analysis is more useful than single-view analysis for image feature extraction.

We can also observe that serial in the sequence of depth-view BNWA, space-view BNWA and pixel-view SimAM (Scheme \#25), which is adopted in our proposed framework, has the best performance overall. We also observe that the overall performances of connecting any two submodules in parallel and then connecting them in series with the third submodule (the mix mode) are significantly worse than that of directly connecting three submodules in series (Scheme \#14 to \#19 vs. Scheme \#20 to \#25). The effect of parallelizing all three submodules is also worse than serializing them (Scheme \#13 vs. Scheme \#20 to \#25). This is consistent with our original design intuition - serial stacking of submodules does bring a deeper network structure than parallelism, and is indeed more conducive to the extraction of multi-level semantic information.

\subsubsection{The effect of VKGD}

\begin{table*}
	\caption{Performance of models using VKGD with or without the attention mechanism. ``VKGD w/o att'' represents the model using VKGD without the attention mechanism. ``VKGD'' represents the model using VKGD. Note that in both models, the GIA module is incorporated in the visual encoder. Thus, VKGD is our proposed model. Best results are in bold.}
	\label{table:ablation_study_VKGD}
	\begin{center}
		\begin{tabular}{c|cccccc}
			\hline
			\multirow{2}{*}{Model} & \multicolumn{6}{c}{NLG Metrics} \\
			& BL-1 & BL-2 & BL-3 & BL-4 & MTR & RG-L  \\
			\hline
			VKGD w/o att & 0.492 & 0.326 & 0.237 & 0.179 & 0.204 & 0.396 \\
			\textbf{VKGD (ours)} & \textbf{0.523} & \textbf{0.357} & \textbf{0.258} & \textbf{0.191} & \textbf{0.226} & \textbf{0.405} \\
			\hline
		\end{tabular}
	\end{center}
\end{table*}

By the comparison of Model-2 and the baseline model, and the comparison of our proposed model IVGN and Model-1 in Table \ref{table:ablation_study}, we find that the proposed VKGD is of great importance. It gains significant improvements in all metrics. This shows that the decoder can adaptively rely on images for text generation. This validates our intuition that by learning the association between the current image and the previously predicted word, the model can better guide the text generator to generate the next word based on the current image.

In addition, comparing our proposed model with the model using VKGD without attention mechanism, as shown in Table \ref{table:ablation_study_VKGD}, we can see that if the attention mechanism is not adopted in VKGD, the performance of the model will be greatly inferior. This proves the contribution of the attention mechanism in our VKGD.

In summary, after the integration of all these submodules, the performance of our model far exceeds the baseline in all metrics, with an average improvement of 8.2\%, among which the highest improvement is the BL-1 metric, with an improvement of 12.7\%.

\subsection{Further analysis}

\subsubsection{Qualitative analysis}

\begin{figure*}
	\centering
	\subfigure{\includegraphics[width=1\columnwidth]{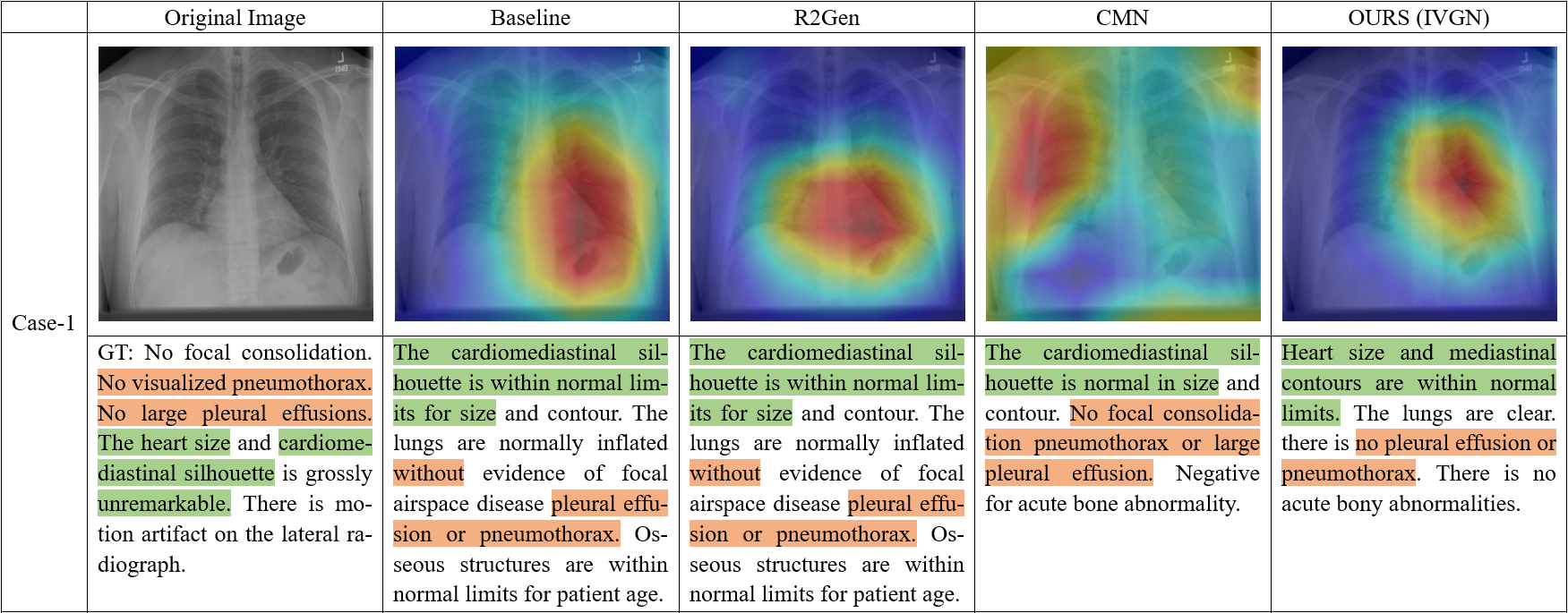}}
	\subfigure{\includegraphics[width=1\columnwidth]{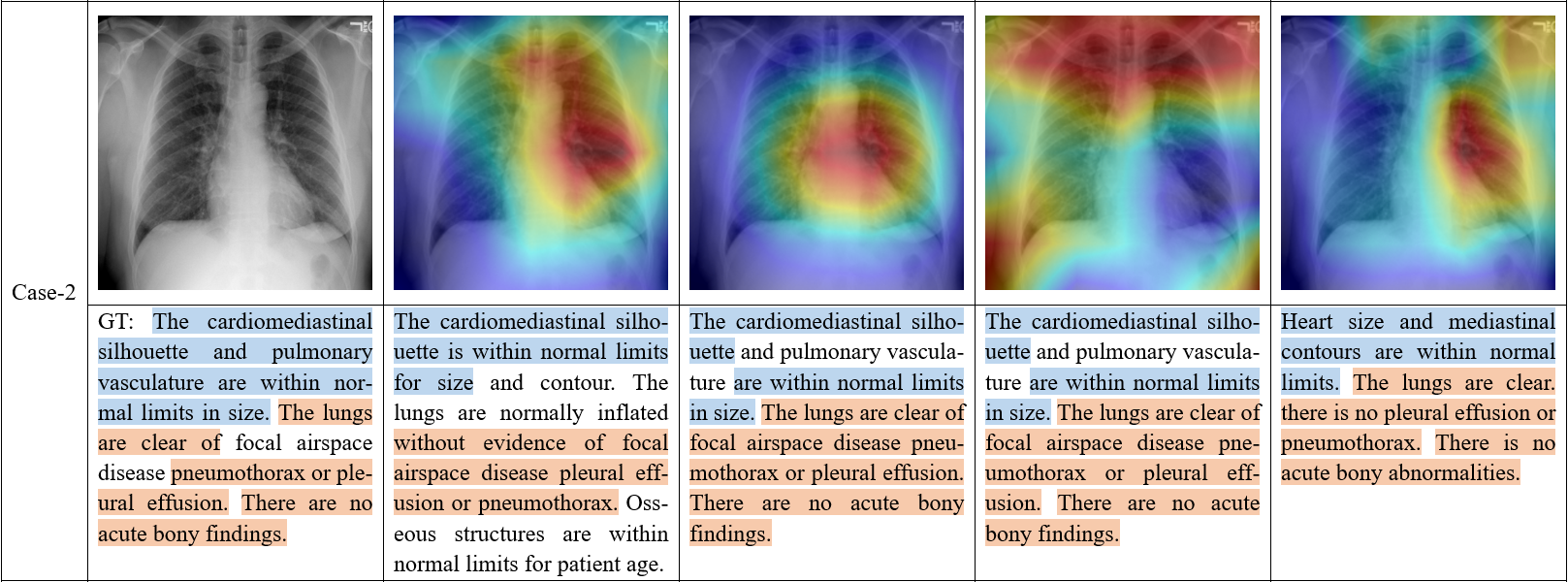}}
	\subfigure{\includegraphics[width=1\columnwidth]{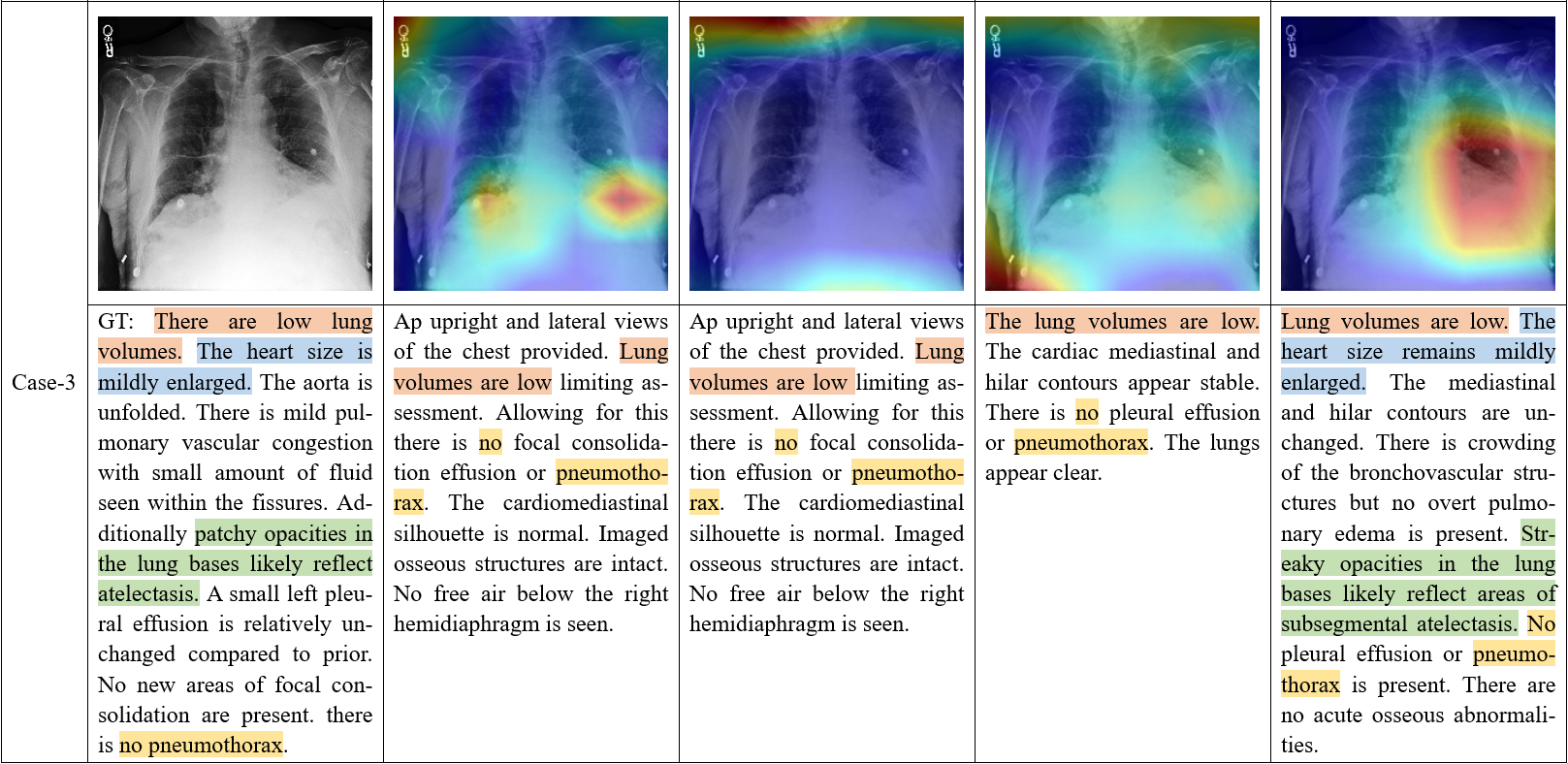}}	
\end{figure*}
\addtocounter{figure}{0}
\begin{figure*}
	\addtocounter{subfigure}{2}
	\centering
	\subfigure{\includegraphics[width=1\columnwidth]{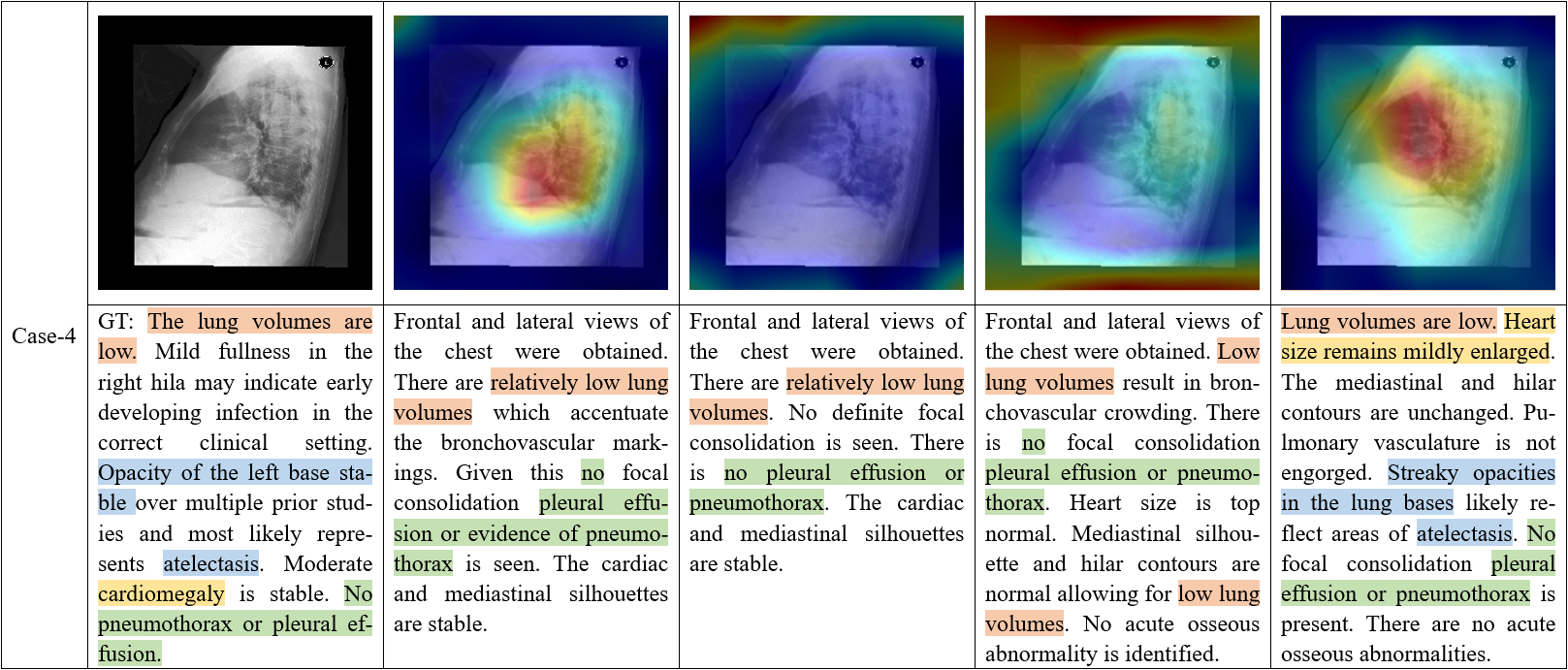}}
	\caption{Visualization of the original images of four cases and their attention heatmaps under several models, including Baseline, R2Gen, CMN and our proposed model, as well as the ground-truth reports and the final reports generated by these models. Case-1 and Case-2 are from \textsc{IU X-Ray}, and Case-3 and Case-4 are from MIMIC-CXR. Text shading in different colors indicates descriptions of different symptoms. GT: Ground-truth report. Best viewed in color.}
	\label{fig_7:vis1}
\end{figure*}

\begin{figure*}
	\centering
	\begin{tabular}{c}
		\includegraphics[width=1\columnwidth]{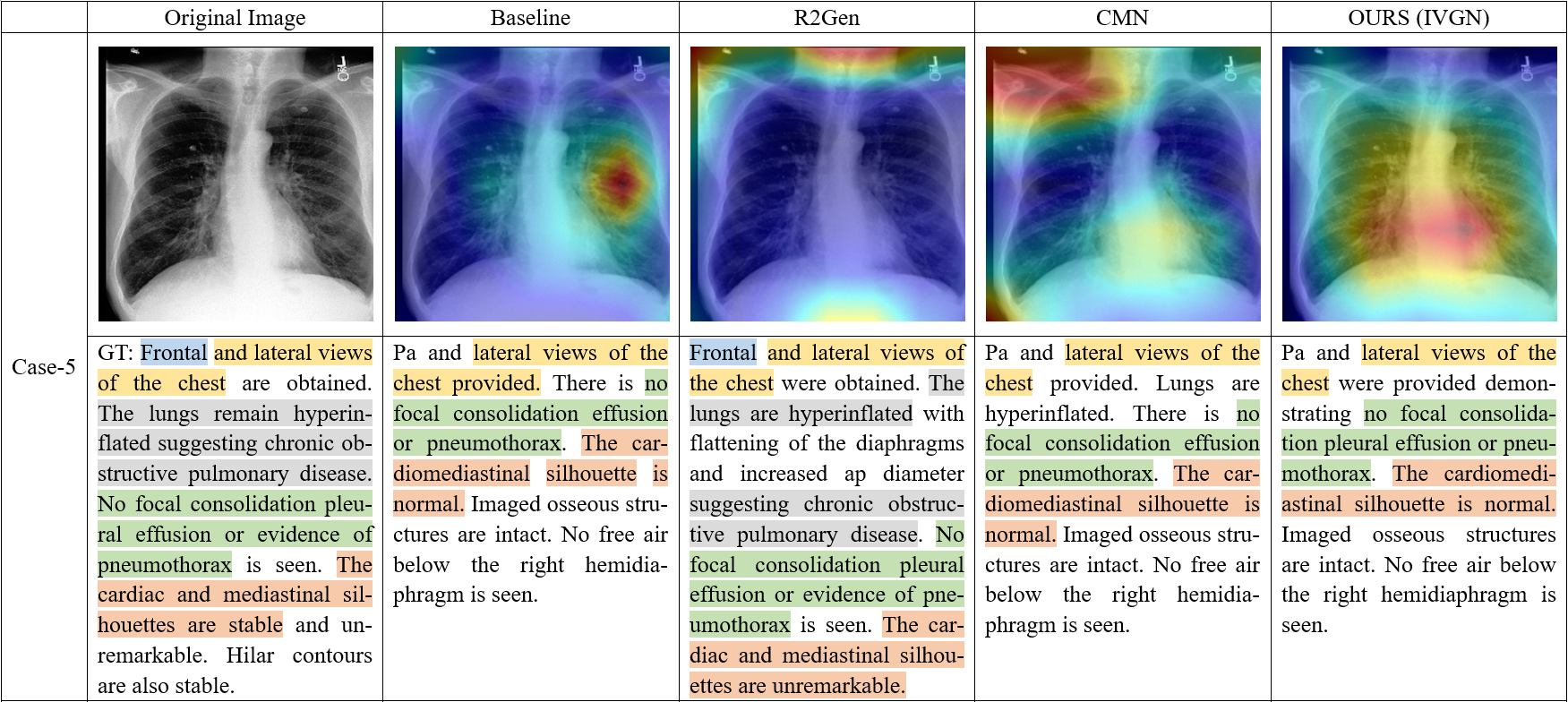} \\
		\includegraphics[width=1\columnwidth]{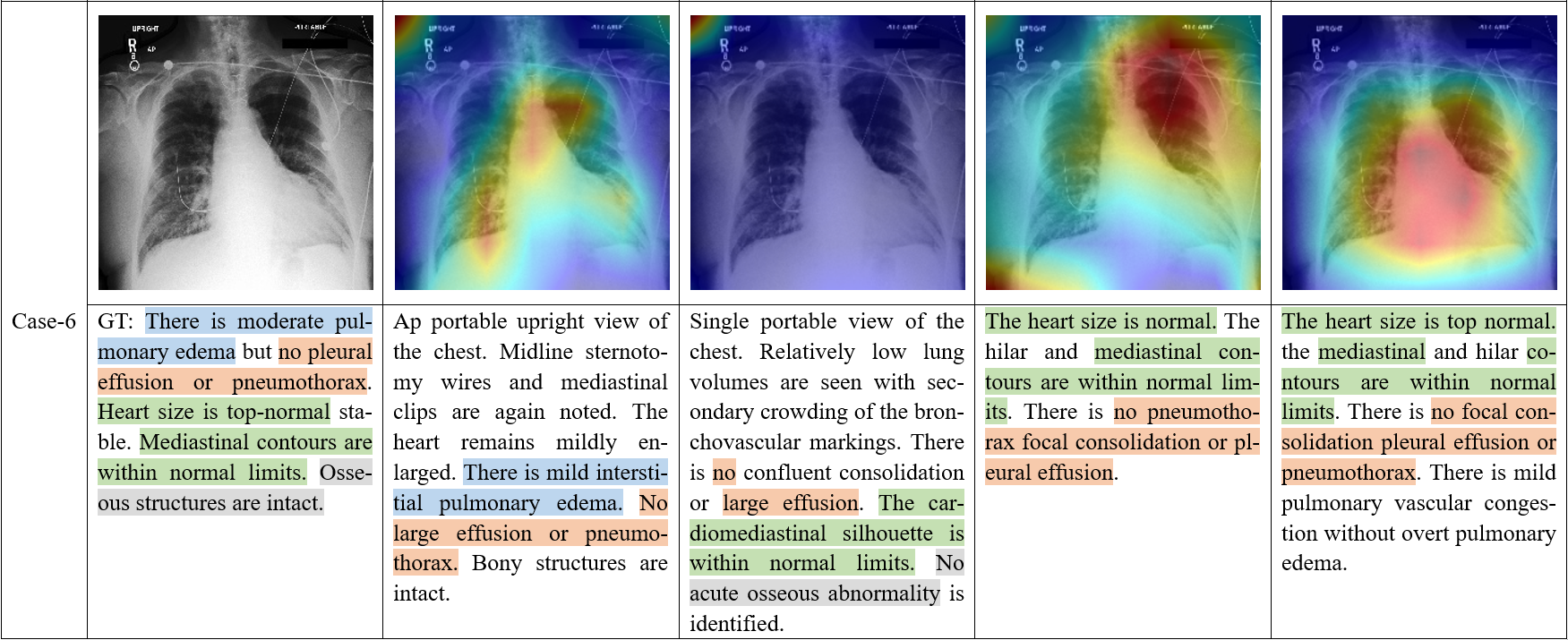} \\
	\end{tabular}
	\caption{Visualization of the original images of two cases where our model generates unsatisfactory reports and their attention heatmaps under several models, including Baseline, R2Gen, CMN and our proposed model, as well as the ground-truth reports and the final reports generated by these models. Both cases are from MIMIC-CXR. Text shading in different colors indicates descriptions of different symptoms. GT: Ground-truth report. Best viewed in color.}
	\label{fig_8:vis2}
\end{figure*}

We visualized and qualitatively analyzed some cases to visually verify the effect of the model, as shown in Figure \ref{fig_7:vis1} and Figure \ref{fig_8:vis2}. For each case, we presented the original image and attention heatmaps under several models, including Baseline, R2Gen, CMN and our proposed model IVGN, and printed the ground-truth report and the final reports generated by these models. Text shading in different colors indicates descriptions of different symptoms. As can be seen from Case-1 to Case-4 in Figure \ref{fig_7:vis1}, the overall effect of our model is superior to other models both in the heatmap of key regions obtained from the images and in the generated report. For example, in Case-3, there are patchy opacities in the lung bases in the image and the ground-truth report shows ``patchy opacities in the lung bases likely reflect atelectasis''. Our model successfully detects this trait in the image, and successfully generates the corresponding diagnostic text ``Streaky opacities in the lung bases likely reflect areas of subsegmental atelectasis'' in the report. However, the reports generated by other models completely ignore this. In addition, the patient's heart is slightly enlarged, which is also detected by our model. Our model manages to focus on the heart (as shown in the heatmap in Colomn ``Case-3'', Row ``OURS (IVGN)'' in Figure \ref{fig_7:vis1}) and generate an accurate diagnosis in the report. Other models still fail to detect this feature in the image and do not describe it in the report. Similarly, in Case-4, the patient is diagnosed as ``most likely represents atelectasis'' and has an enlarged heart ``cardiomegaly''. Our model accurately detects and describes these two phenomena, while other models completely ignore these two important image features and therefore cannot accurately describe them.

\subsubsection{Error analysis} We also selected two cases where our model generates unsatisfactory reports for analysis, as shown in Case-5 and Case-6 in Figure \ref{fig_8:vis2}. In Case-5, the patient's lungs are ``hyperinflated suggesting chronic obstructive pulmonary disease''. Our model does not discover this problem (while R2Gen successfully discovers this problem but the baseline model and CMN also neglect it). The possible reason is that our model focuses on local areas such as the interior of the thorax, mediastinal of the heart, while ignoring the contour of the thorax. Similarly, in Case-6, our model ignores osseous structure, which also belongs to contour features. To solve this problem, we will focus on how to make the model extract more accurate contour features in the subsequent research. Despite this, the model still performs well in terms of attention heatmap and the rest of the report generation in both cases.

\subsubsection{NLG metrics analysis}

We observed that the METEOR (MTR) and ROUGE-L (RG-L) metrics of the model are not optimal on MIMIC-CXR. Therefore, we visualized some reports generated by our model and compared them with the ground-truth reports for further analysis. We found that:
	
First, the disease symptoms may not appear in the same order in the report generated by our model as in the ground-truth report. For example, in Case-4 of Figure \ref{fig_7:vis1}, the description in the ground-truth report is ``No \underline{pneumothorax} or \underline{pleural effusion}.'', whereas the description in our generated report is ``No focal consolidation \underline{pleural effusion} or \underline{pneumothorax} is present.'' It can be seen that the prediction of diagnosis is very accurate, but the inconsistency of ``pleural effusion'' and ``pneumothorax'' in the generated report and the ground-truth report leads to the relatively low METEOR and ROUGE-L scores. This is because that one of the concerns of the METEOR metric is the length of consecutive words, and the calculation of ROUGE-L metric depends on the longest common subsequence, which is greatly affected by the order of words. By simply swapping the positions of ``pleural effusion'' and ``pneumothorax'' in our generated report, the METEOR score of our generated report increased from 0.156 to 0.161, and the ROUGE-L score increased from 0.290 to 0.329.

Second, the sequence of the sentences describing different diseases and symptoms in the generated report may be different from that in the ground-truth report. Take Case-6 in Figure \ref{fig_8:vis2} as an example, we can see that the sequence of the sentences describing different diseases and symptoms are indeed different in our generated report and in the ground-truth report (sentences with light green and orange shading). However, these sentences are independent of each other, and the sequence in which they appear has no effect on the diagnosis of the patient's condition. By simply switching the sequence of these two sentences, the ROUGE-L score of our generated report is increased from 0.361 to 0.601. In this example, the METEOR score is not affected by the sentence sequence adjustment, because the METEOR metric only looks at a limited number of consecutive words.

\subsubsection{CE metrics analysis}

We noticed that the precision, recall and F1-score of all models are relatively low compared to that of ordinary classification tasks. Thus, we further explored the calculation method of the CE metrics and analyzed some visualized cases. The overall precision, recall and F1-score for a particular patient's report is the average of the precision, recall and F1-score over the 14 categories. By visualizing the ground-truth report, we observe that the ground-truth reports of most patients cover only a few categories of diseases, which leads to a precision, recall and F1-score of 0 for many categories of the patient and finally results in the relatively low overall precision, recall and F1-score. We further observed the reports in case 3 to case 6 in Figure \ref{fig_7:vis1} and Figure \ref{fig_8:vis2}. It can be seen that on the whole, the reports generated by our model in these cases are already very close to the ground-truth reports. However, their CE metrics are still relatively low, with a precision of 0.488, a recall of 0.536, and a F1-score of 0.505.

\subsection{Analysis of parameters and FLOPs}

\begin{figure}
	\centerline{\includegraphics[width=0.4\columnwidth]{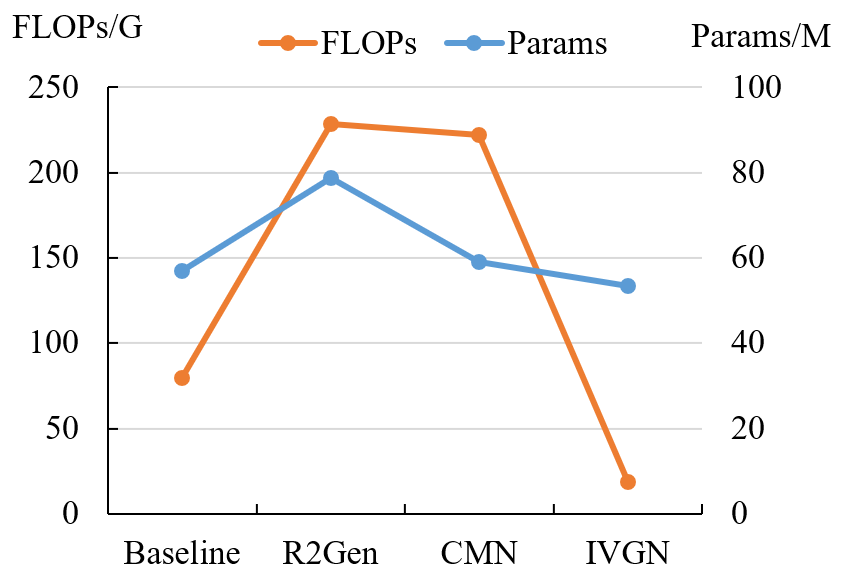}}
	\caption{The analysis of parameters and FLOPs of Baseline, R2Gen, CMN and our model (IVGN). Best viewed in color.}
	\label{fig_9}
\end{figure}

We also analyzed the number of parameters and floating-point operations per second (FLOPs) of Baseline, R2Gen, CMN and our model. We did not compare the parameters and operations of all models mentioned in Table \ref{table:comparisons_with_previous} because details of the parameters of many models were not published and the original codes of some models were not available. As can be seen from Figure \ref{fig_9}, the number of parameters and FLOPs of our model is significantly lower than that of other models, which shows that our model is lightweight and has the potential to be deployed on a variety of terminals.

\subsection{Limitations}

This study has several limitations. First of all, it is foreseeable that the model may not perform well on medical images of other modalities (e.g. MRI, CT, ultrasound), or reports in other languages (e.g. Chinese, Spanish) that it has never seen before. This is intuitive because the model has only been trained on datasets of X-ray images with reports in English, and there are significant differences between different medical imaging modalities and natural languages. In future research, we will consider training the model on more diverse data so that it can learn the differences between different modalities and languages, and expanding the model to be one that can cope with a wide variety of image modalities and report languages. Secondly, when applying this model to unseen X-ray images in real-world clinical scenarios, the robustness and generalization ability of the model may also not be good enough due to objective factors such as different X-ray acquisition instruments, different disease types, and different report writing styles of radiologists (detailed in Section S3 in the supplementary material). In future work, we will explore how to align and normalize data from different sources to alleviate data bias caused by different data acquisition devices and writing styles, and explore how to make better use of data from multi-center to train the model into a robust and generalizable model. In addition, the model adopts a visual extractor pre-trained on ImageNet, and the natural difference between natural images and medical images result in limited effectiveness of the pre-training. In this study, we just made some cursory explorations about pre-training (detailed in Section S1 in the supplementary material). In future studies, we will explore more comprehensively whether pre-training based on medical image datasets can improve model performance or accelerate model convergence. Leveraging the transferrable representation learning capabilities of large-scale visual language pre-trained models could be a potential research direction.

\section{Conclusion}

Automatic radiology report generation using AI model requires that the model can not only correctly detect the organs or lesions in the image, but also understand the relationship between them, and finally describe their location and relationship with clinically accurate, reasonable and fluent language. The existing models are unsatisfactory in both image feature extraction and text generation. On the one hand, they do not comprehensively consider multi-view visual information, which is particularly important for radiology image feature extraction. On the other hand, in report generation, they fail to make full use of image features to guide text generation, either deviating from image features or relying too much on image features. To improve the above two problems, we propose an Intensive Vision-guided Network (IVGN), which contains a Globally-intensive Attention-guided Visual Encoder and a Visual Knowledge-guided Decoder (VKGD). In the Globally-intensive Attention-guided Visual Encoder, a lightweight Globally-intensive Attention (GIA) module is proposed to better mine salient image features from multiple perspectives of the radiology images. And the VKGD focuses on excavating the relationship between the previously predicted words and the current image, and adaptively uses image features to guide report generation. Extensive experiments on two commonly-used benchmark datasets \textsc{IU X-Ray} and MIMIC-CXR show that our model has higher report accuracy than existing models, and can generate more enlightening attention heatmaps with significantly lower number of parameters and FLOPs. In subsequent research, we will focus on how to extract more accurate global and contour features from radiology images, so as to better generate text description of global and contour information in reports. In addition, we will explore how to use multi-center data training and large-scale visual language pre-training models to improve the robustness and generalization ability of the model.

\section*{Acknowledgments}

This work was supported in part by the National Natural Science Foundation of China (Grant No. U1811461), the Key Areas Research and Development Program of Guangdong (Grant No. 2018B010109006), and Guangdong Introducing Innovative and Entrepreneurial Teams Program (Grant No. 2016ZT06D211).

\section*{Conflict of Interest Statement}

We have no conflicts of interest to disclose.

%

\end{document}